\documentclass[times,review,10pt]{elsarticle}
\usepackage{longtable}
\usepackage{amsmath}
\usepackage{booktabs}
\usepackage{subcaption}
\usepackage{bbm}
\usepackage{setspace}
\usepackage{xcolor}
\usepackage{environ}
\journal{Pattern Recognition}
\usepackage[breaklinks,hidelinks]{hyperref}
\usepackage{amsmath,amsfonts}
\usepackage{algorithmic}
\usepackage{algorithm}
\usepackage{array}
\usepackage{booktabs}
\usepackage{textcomp}
\usepackage{stfloats}
\usepackage{url}
\usepackage{verbatim}
\usepackage{graphicx}
\usepackage{multirow}

\usepackage{hyperref}
\usepackage{arydshln}
\usepackage{color}
\newcommand{\First}{\textcolor{red}}
\definecolor{MyDeepGreen}{RGB}{0,150,0} 
\newcommand{\Second}{\textcolor{MyDeepGreen}}
\definecolor{MyDeepBlue}{RGB}{50,150,240} 
\newcommand{\Third}{\textcolor{MyDeepBlue}}

\usepackage{tikz}

\NewEnviron{commentD}{%
    \color{black}
    \BODY
}

\begin{document}

\begin{highlights}
\item{}%
A large-scale SOD dataset that incorporates various weather noise
\item{}%
A dual-branch network for salient object detection in adverse weather conditions
\item{}%
A benchmark for RGB SOD in adverse weather conditions
\end{highlights}

\begin{frontmatter}

\title{WXSOD: A Benchmark for Robust Salient Object Detection in Adverse Weather Conditions}

\author[1]{Quan Chen}
\ead{chenquan@alu.hdu.edu.cn}
 
\author[2]{Xiong Yang}
\ead{22080839@hdu.edu.cn}

\author[3]{Bolun Zheng\corref{cor1}}
\ead{blzheng@hdu.edu.cn}

\author[3]{Rongfeng Lu}
\ead{rongfeng-lu@hdu.edu.cn}

\author[2]{Xiaokai Yang}
\ead{242080082@hdu.edu.cn}

\author[3]{Qianyu Zhang}
\ead{zqqqyu@163.com}

\author[4]{Yu Liu}
\ead{liuyu77360132@126.com}

\author[3]{Xiaofei Zhou}
\ead{zxforchid@outlook.com}

\affiliation[1]{organization={Department of Artificial Intelligence},
addressline={Jiaxing University}, 
city={Jiaxing},
postcode={314001}, 
country={China}}

\affiliation[2]{organization={School of Communication Engineering}, 
addressline={Hangzhou Dianzi University}, 
city={Hangzhou},
postcode={310018}, 
country={China}}

\affiliation[3]{organization={School of Automation},
addressline={Hangzhou Dianzi University}, 
city={Hangzhou},
postcode={310018}, 
country={China}}

\affiliation[4]{organization={Department of Electronic Engineering},
addressline={Tsinghua Uni
versity}, 
city={Beijing},
postcode={100084}, 
country={China}}

\cortext[cor1]{Corresponding Author}


\begin{abstract}
Salient object detection~(SOD) in complex environments remains a challenging research topic. Most existing methods perform well in natural scenes with negligible noise, and tend to leverage multi-modal information (\textit{e.g.}, depth and infrared) to enhance accuracy. However, few studies are concerned with the damage of weather noise on SOD performance due to the lack of dataset with pixel-wise annotations. To bridge this gap, this paper introduces a novel Weather-eXtended Salient Object Detection (WXSOD) dataset. It consists of 14,945 RGB images with diverse weather noise, along with the corresponding ground truth annotations and weather labels. To verify algorithm generalization, WXSOD contains two test sets, \textit{i.e.}, a synthesized test set and a real test set. The former is generated by adding weather noise to clean images, while the latter contains real-world weather noise. Based on WXSOD, we propose an efficient baseline, termed Weather-aware Feature Aggregation Network~(WFANet), which adopts a fully supervised two-branch architecture. Specifically, the weather prediction branch mines noise-related deep features, while the saliency detection branch fuses semantic features extracted from the backbone with noise-related features for SOD. Comprehensive comparisons against 17 SOD methods shows that our WFANet achieves superior performance on WXSOD. The code and benchmark results are publicly available at \url{https://github.com/C-water/WXSOD}
\end{abstract}


\end{frontmatter}

\section{Introduction}\label{sec:introduction}
Salient object detection attempts to mimic the human visual system and precisely highlights the most attractive regions in an images~\cite{cong2018review}.
Advances in deep learning and diverse datasets continue to drive the field of salient object detection. 
Numerous works~\cite{wei2020label,fang2021ibnet,feng2020residual,wu2019cascaded,li2018contour} are emerged to overcome the problem of inaccurate segmentation due to the non-uniform size and complex details of targets.
However, few methods are concerned with the damage of various weather noise on SOD performance.

\begin{figure}[!h]
\centering
\includegraphics[width=1.0\linewidth]{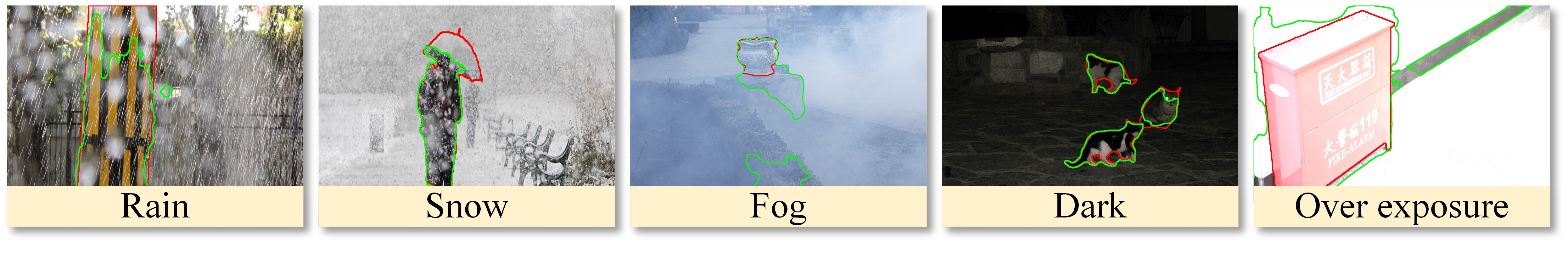}
\caption{Common weather noise interferes with salient objects. Green curve represents the edge of the salient target predicted by MEANet\cite{liang2024meanet}, while red curve represents the ground-truth.}
\label{Weather disturbance}
\end{figure}  

Images captured in the real world inevitably contain weather-related noise, such as rain, snow, fog, low-light, and over-exposure. These degradations pose challenges to existing SOD methods. As shown in Fig.~\ref{Weather disturbance}, different types of weather noise interfere with algorithms in distinct ways.
For instance, rain and snow are scattered throughout the image, thus destroying local details of salient objects; fog leads to loss of information over large areas; while extreme lighting conditions reduce content contrast, which directly weakening target salience. To improve the detection accuracy, one feasible strategy is to employ image restoration models~\cite{zamir2022restormer} to reconstruct a clean version of the noise image, and then predict salient objects.
Similar solutions have been proved to be effective in object detection~\cite{jiang2023dawn} and semantic segmentation~\cite{sun2023event}, but only for a single type of weather noise.
Several image restoration efforts~\cite{li2022all,potlapalli2024promptir,ai2024multimodal} attempt to design a unified framework to remove various weather noises, which seems to provide a convenient solution for the salient target detection in adverse weather conditions.
However, such two-stage solutions suffer critical drawbacks: 1) Unreliable image restoration models may corrupt RGB images, thus reducing detection accuracy; 2) image pre-processing brings additional computational costs, resulting in lower detection efficiency.
Consequently, a one-stage SOD algorithm tailored to combat weather noise is preferable, with two key factors: data and feature extraction. The lack of datasets restricts the research of SOD in adverse weather conditions.
Most existing SOD datasets~\cite{yang2013saliency,wang2017learning,li2014secrets,li2019nested,zhang2020dense} consist of clean images, inherently limiting the performance of learning-based methods on noisy inputs. 
Only a handful of datasets~\cite{tu2022rgbt,wang2024alignment,song2023modality} explicitly incorporate extreme illumination conditions into their image samples, partially addressing the mismatch between existing datasets and real-world complex environments.
Since various weather noise show different effects on image salience, we argue that another key is to mine weather-related representations to optimize SOD.
Previous data-driven SOD methods focus on single-modal feature extraction or multi-modal feature interaction, but their architectures have no branches for learning noise representations with distinct characteristics.

In this paper, we construct a comprehensive benchmark for RGB SOD in adverse weather conditions. 
To alleviate the limited data issue, we introduce WXSOD, which incorporates diverse weather-induced noise.
WXSOD comprises 14,945 RGB images with pixel-wise annotations, covering over 5,000 scenes, and is divided into a synthesized training set, a synthesized test set and a real test set. For synthesized data, we leverage an image-based style transformation library~\cite{jung2020imgaug} to inject noise into clean images, thus simulating extreme weather environments. To enhance the realism of WXSOD, we also collect 554 images with real weather noise.

Secondly, we propose a simple yet effective baseline termed Weather-aware Feature Aggregation Network (WFANet), which comprises a saliency detection branch~(Branch-1) and a weather prediction branch~(Branch-2). We argue that explicitly learning noise-related features under diverse weather conditions facilitates the mining of salient features. Specifically, the weather prediction branch leverages a classification task as a proxy to learn noise-related features: a backbone network maps images into high-level features, and a classifier is then introduced to predict weather noise categories. The input features to this classifier can be regarded as noise-specific feature representations.
Next, the saliency detection branch employs a backbone network to extract multi-scale semantic features, refined through cross fusion module~(CFM) that integrate features of adjacent scales. Subsequently, multi-source fusion modules (MSFM) integrate noise-related features with semantic features to construct unified feature representations. Finally, the output features from each MSFM are concatenated for salient object prediction. Comprehensive evaluations on WXSOD demonstrate that compared with 17 RGB SOD methods, WFANet achieves superior segmentation accuracy on both the synthesized test set and the real test set. 

In summary, the main contributions of this paper include:
\begin{itemize}
    \item We introduce a large-scale dataset named WXSOD, designed to facilitate robust salient object detection under adverse weather conditions. WXSOD comprises 14,945 RGB images with diverse weather noise, along with ground-truth annotations and weather category labels.
    \item We propose the dual-branch network, WFANet, where a weather prediction branch employs classification as a proxy to learn noise-related feature. These features are fed into the saliency prediction branch for joint salient object prediction.
    \item Building upon WXSOD, we establish the first benchmark for RGB SOD under adverse weather conditions, which includes 17 comparison methods. Experimental results show that WFANet can accurately segment salient objects from images corrupted by diverse weather noises.
\end{itemize}

The rest of this paper is structured as follows. We first introduce related works in Section~\ref{sec:Related Works}. In Section~\ref{sec:WXSOD Benchmark}, we present WXSOD dataset in detail.
Then, we introduce the network WFANet and its key modules in Section~\ref{sec:Methods}.
The experimental results and analysis are presented in Section~\ref{sec:Experiments}.
Finally, the concluding remarks are drawn in Section~\ref{sec:Conclusion}.

\section{Related Works}\label{sec:Related Works}

\subsection{SOD Datasets}
Over the past dozen years, plentiful datasets have been built to evaluate SOD methods. According to the modal type of inputs, existing datasets can be divided into four categories: \textit{RGB}, \textit{RGB-D}, \textit{RGB-T} and \textit{RGB-DT} datasets. As the name implies, the \textit{RGB} datasets contain images captured by pixel sensors and corresponding labels. In practice, contour maps of targets are usually drawn as auxiliary features to enhance edge details of predicted regions. Except for ground-view images, partial \textit{RGB} datasets also cover remote sensing images~\cite{li2019nested,zhang2020dense} and underwater image~\cite{jian2017ouc,islam2022svam,hong2023usod10k}. With the popularity of visual sensors, several works attempted to leverage multi-modal information to improve detection performance, so they captured depth and thermal maps while acquiring RGB images to build \textit{RGB-D}~\cite{peng2014rgbd,cheng2014depth,zhu2017three,niu2012leveraging} and \textit{RGB-T}~\cite{wang2018rgb,tu2019rgb,tu2022rgbt} datasets. To mine the complementary information of various modalities, Song~\textit{et al.}~\cite{song2022novel} presented a novel \textit{RGB-DT} dataset consisting of triple-modal images (\textit{i.e.}, visible image, depth image, and thermal image). The YLLSOD~\cite{yu2024degradation} is dedicated to SOD in low-light scenes, while the VDT5000~\cite{tu2022rgbt} introduced some low light and over-exposure images. 
Although the UAV RGB-T 2400~\cite{song2023modality} dataset incorporated scenes with dark, overexposure, and fog conditions, it is limited by an incomplete range of weather noises and a failure to distinguish between different categories.
To promote SOD research in adverse weather conditions, we create a novel dataset WXSOD containing a variety of weather noises, such as rain, snow, fog, low light, over-exposure, rain+snow, rain+fog and snow+fog.

\subsection{SOD Methods}
Salient object detection aims to detect the visual attractive regions from images or videos.
Traditional algorithms mainly rely on the hand-crafted features~\cite{achanta2009frequency,peng2016salient,zhang2015minimum}, which are incapable of coping with complex scenes due to poor feature representations.
Drawing from the success of deep learning in numerous vision tasks~\cite{chen2024sdpl,zheng2021learning}, researchers have also attempted to design various learning-based saliency models.
Depending on the type of inputs, the above models are generally divided into two categories: single-modal SOD and multi-modal SOD.

\textbf{1) Single-modal SOD.} 
Earlier methods~\cite{wei2020label,fang2021ibnet,feng2020residual,he2025samba} only take RGB images as input, and design numerous effective network framework.
Several typical structures, \textit{e.g.}, residual connection~\cite{he2016deep}, have been shown to improve salience prediction performance and continue to inspire subsequent researches.
For example, Chen~\textit{et al.}~\cite{chen2018reverse} designed reverse attention to enhance features and filter-out backgrounds.
In addition, edge prior information plays a crucial role in capturing intricate details within boundary regions, so some studies~\cite{song2023salient,wu2024misclassification,wu2025weakly,li2025ifa,zhu2025dc} introduce edge-aware learning to assist the saliency prediction.
A simple yet effective strategy is to employ an additional edge-aware loss, \textit{i.e.}, calculating the gradient of an RGB image to aid the learning of boundary details.
Besides, several works~\cite{liu2020dynamic,zhou2020interactive} construct multi-task learning frameworks to learn the edge information through an auxiliary edge-aware branch.
To mitigate the influence of dark degradation and low contrast, Yu~\textit{et al.}~\cite{yu2024degradation} combined the SOD network with a low-light image enhancement method to form a novel learning framework.

\textbf{2) Multi-modal SOD.}
As complementary information to RGB images, depth and thermal are widely used in the SOD task and show unique advantages.
Depth maps contain the relative spatial distances of objects, which facilitates the segmentation of salient targets from adjacent regions in color space.
To explore the correlations and differences between RGB features and depths, plentiful cross-modal learning frameworks have been proposed~\cite{wen2021dynamic,zhu2024cmignet,yang2025mitigating,zhong2025lesod} and promoted a novel SOD task, called \textit{RGB-D} SOD.
For instance, Gao~\textit{et al.}~\cite{gao2023depth} proposed a depth-aware inverted refinement network to preserve the different level details with multi-modal cues.
However, in scenarios with weak depth variation, depth maps offer limited gain for saliency prediction, especially when the target is coupled to multiple objects at the same depth.
Consequently, several works~\cite{cong2022does,zhou2024frequency,wang2024learning,zhou2025deformation} explore thermal priors, which can distinguish adjacent objects based on object temperature and are robust to changes in illumination.
To show advantages of thermal maps in complex scenes, Tu~\textit{et al.}~\cite{tu2022rgbt} collected a large-scale \textit{RGB-T} dataset with degraded samples, and proposed a end-to-end network to adaptively select salient cues from two modal features.
To further enhance the performance of saliency detection, researchers~\cite{luo2024dynamic,wan2024tmnet,bao2024quality,wang2025unified} pay attention to fusing three modal data and introduce a new task, namely visible-depth-thermal~(VDT) SOD.

However, most existing SOD methods are suitable for normal environments and lack of consideration for common weather noise in real world. While depth and thermal maps can provide complementary information for RGB images, low-quality modalities obtained in adverse environments, such as rain and snow, are equally noisy. In this paper, we design a unified framework to deal with SOD in various weather conditions.

\section{WXSOD Benchmark}\label{sec:WXSOD Benchmark}
In this section, we first elaborate on the differences between the WXSOD dataset and existing datasets. 
Subsequently, we introduce the key features of the WXSOD dataset and corresponding build details.
Then, we provide some synthesized/real samples in Fig.~\ref{Visualization of different content} and Fig.~\ref{Visualization of synthesized weather noise}, and statistical analysis in Table~\ref{Statistical analysis of size and number} and Table~\ref{Statistical analysis images numbers in WXSOD}. Finally, Fig.~\ref{t-sne} illustrates the domain differences between WXSOD and representative datasets.

\begin{table}[!h]
\centering
\caption{Statistics comparison with existing SOD datasets. Noise diversity levels: $\circ$=negligible, $\triangle$=limited, $\square$=moderate,	$\blacksquare$=rich.}
\label{Statistics comparison with existing SOD datasets}
\resizebox{1.0\linewidth}{!}{
\begin{tabular}{cccccccccc}
\hline
\multirow{2}{*}{Datasets} & \multirow{2}{*}{Year} & \multirow{2}{*}{Modality} & Scene     & \multirow{2}{*}{View} & Average & \multicolumn{2}{c}{Noise diversity} & \multicolumn{2}{c}{Image number} \\ \cline{7-10} 
&                       &                           & Number       &    & Resolution & Synthetic      & Real     & Training set      & Test set     \\ \hline
DUT-O~\cite{yang2013saliency} &   2013    &  RGB    &  5168   &   Ground  &  376$\times$320    &  $\circ$      &   $\circ$       &   0   &     5168     \\
ECSSD~\cite{shi2015hierarchical} &   2015    &   RGB    &   1000    & Ground  &   375$\times$311   &   $\circ$    &   $\circ$       &    0  &   1000 \\
HKU-IS~\cite{li2015visual} &   2015  &  RGB  &  4447   &   Ground  &   386$\times$292   &  $\circ$              &    $\circ$      &   3000    &   1447  \\
DUTs~\cite{wang2017learning} &   2017   & RGB &  15572    &  Ground  &   377$\times$322     &   $\circ$           &   $\circ$       &  10553   &  5019     \\
ORSSD~\cite{li2019nested} &   2019   &   RGB   &   800   &    Satellite    &  479$\times$426   &    $\circ$     &   $\circ$       &     600   &     200   \\
EORSSD~\cite{zhang2020dense} &   2020   &  RGB      &  2000  &     Satellite   &  523$\times$456  &  $\circ$    &  $\circ$        &   1400   &    600  \\
SIP~\cite{fan2020rethinking} &   2020   &    RGB+D    & 929  &     Ground    &   778$\times$957    &  $\circ$     &   $\circ$       &    0    &   929 \\
VT5000~\cite{tu2022rgbt} & 2022  &  RGB+T  & 5000 &         Ground  &  640$\times$480  &  $\circ$   &      $\triangle$    &     2500   &   2500  \\
UVT2000~\cite{wang2024alignment} & 2024   &  RGB+T   & 2000  &    Ground   & 2048$\times$1536  & $\circ$      &   $\triangle$       &  0  &   2000   \\
UAV RGB-T 2400~\cite{song2023modality}   & 2023     & RGB+T    & 2400    & Drone     & 1920$\times$1080       &   $\circ$     & 	$\square$   & 1200   & 1200    \\ 
VDT2048~\cite{song2022novel}   & 2023   & RGB+D+T   &  2048    & Ground     &   640$\times$480    & $\circ$       & $\triangle$   &  1048  &   1000  \\ 
\hdashline
WXSOD    & 2025    & RGB     & 5554    & Ground   & 580$\times$453       &   $\blacksquare$    &    $\blacksquare$  & 12891    & 2054         \\ \hline
\end{tabular}
}
\end{table}

\subsection{Dataset Overview}
\begin{figure}[!h]
\centering
\includegraphics[width=0.9\linewidth]{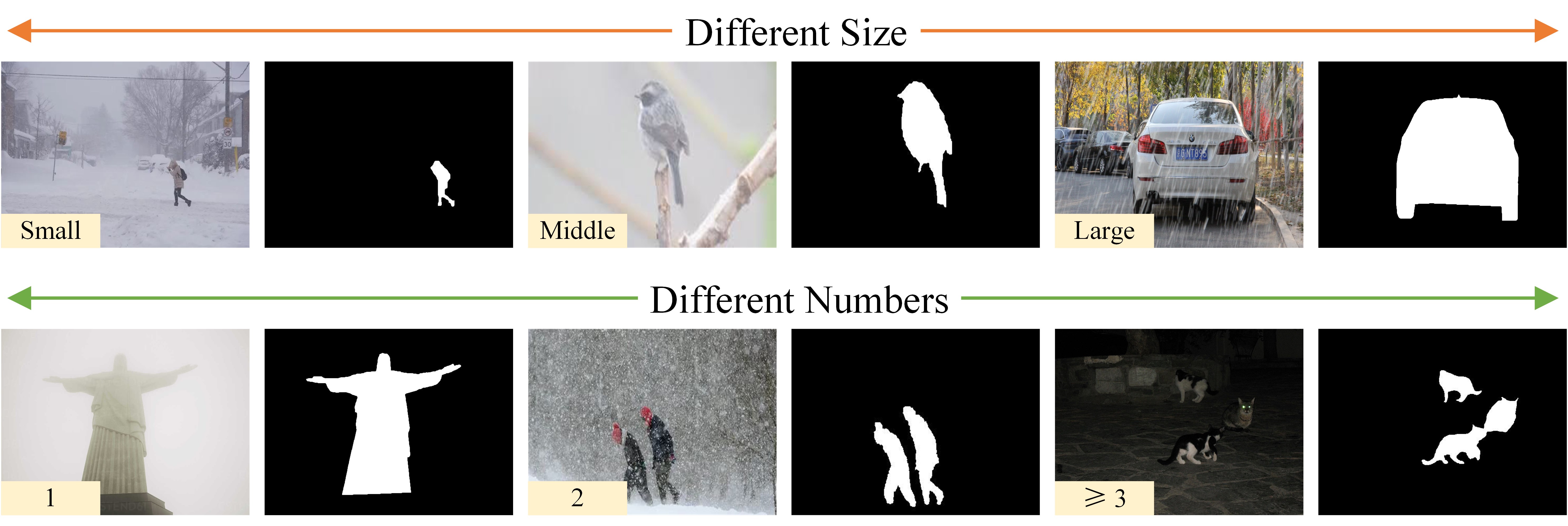}
\caption{Examples of WXSOD dataset. The first row shows scenes with different sizes of salient objects, and the second row shows scenes with different numbers of salient objects.}
\label{Visualization of different content}
\end{figure}

Diverse datasets have contributed to the rapid development of SOD field. However, RGB images of existing datasets are relatively clean, which cause algorithms meet a large performance drops under noisy conditions, especially weather noise conditions~(\textit{e.g}, rain, fog and overexposure). The underlying reason lies in the domain differences between training data and test images. To facilitate the research of SOD under adverse weather conditions, we present a novel dataset, named WXSOD, which explicitly considers domain differences caused by weather noise. As shown in Table~\ref{Statistics comparison with existing SOD datasets}, our WXSOD dataset covers diverse scene types, with the total number of samples exceeding that of most existing datasets. Most importantly, it includes 8 types of synthetic weather noise and 5 types of real weather noise, which better meets the requirements of this study. Its construction details and data-level characteristics are introduced below:

\textbf{1) Diverse detection content.}
To reduce labeling costs and ensure scene diversity, most scenes in WXSOD are taken from existing datasets with clean images and corresponding annotations, including DUTS~\cite{wang2017learning}, DUT-O~\cite{yang2013saliency}, ECSSD~\cite{shi2015hierarchical}, HKU-IS~\cite{li2015visual} and VT5000~\cite{tu2022rgbt}.
First, we retain outdoor scenes compatible with adverse weather noise from the above five datasets. Then, the retained images are randomly divided into two parts according to the ratio of 7:3, which are used as the basic images for the subsequent synthesized training set and synthesized test set respectively.
Benefiting from differences between underlying datasets, RGB images of WXSOD have salient objects of different categories, sizes and numbers, as shown in Fig.~\ref{Visualization of different content}.

\begin{figure}[!h]
\centering
\includegraphics[width=1.0\linewidth]{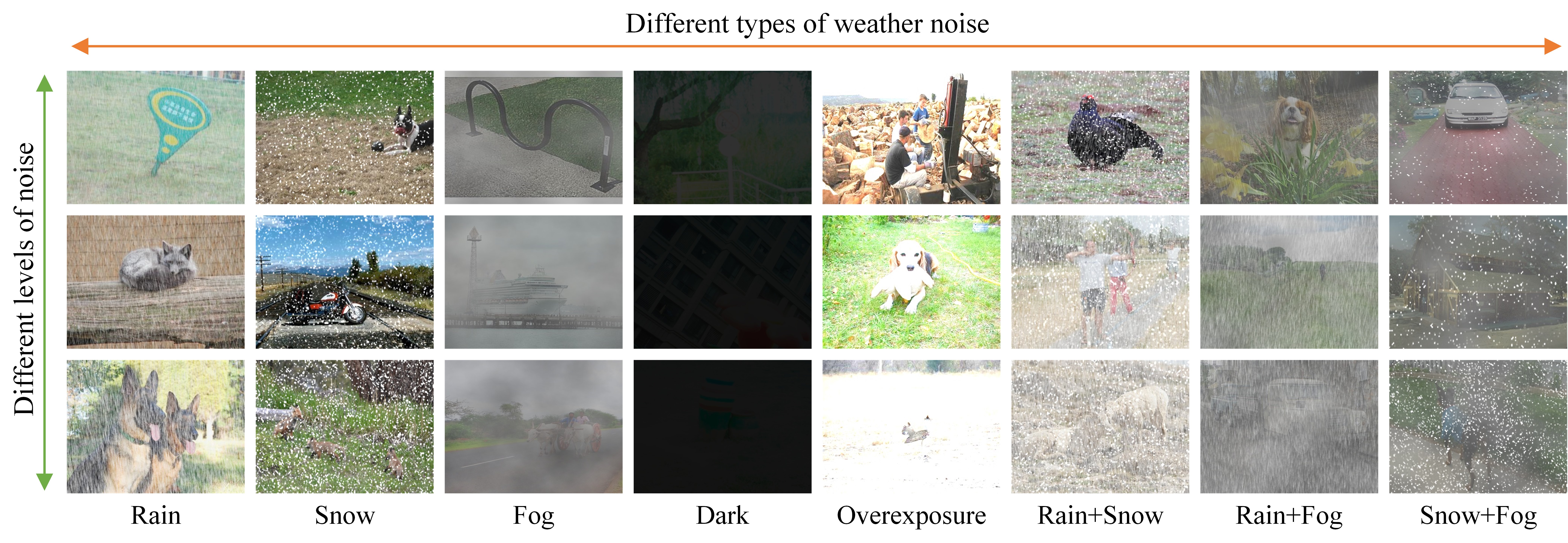}
\caption{Examples of synthesized test set~(a) and real test set~(b) in the WXSOD dataset.}
\label{Visualization of synthesized weather noise}
\end{figure}  

\textbf{2) Different weather noise.}
Given the high costs of collecting real-world data, numerous vision tasks~\cite{ji2022information,ji2023hyformer,wang2024multiple} tend to generate synthetic realistic noise data for training, and achieve satisfactory results.
Inspired by this, we choose an off-the-shelf image-based style transformation library~\cite{jung2020imgaug} to simulate weather noise, including fog, rain, snow, dark, overexposure, fog+rain, fog+snow and rain+snow~(see Fig.~\ref{Visualization of synthesized weather noise}).
For synthesized training set, each RGB image is randomly processed into 2-5 degraded images with various noise types, while the original image is randomly retained.
For synthesized test set, each RGB image is unique, \textit{i.e.}, either one noise-specific or noise-free image.
Besides, each type of weather noise contains multiple levels of intensity.
The complex noise types, as well as the variety of noise intensity, greatly increase the challenge of the proposed WXSOD.

\textbf{3) Real scene.}
There is domain shift between clean and noise images, and also between synthetic noise and real noise. To evaluate the generalization of WXSOD-based methods, we collect 554 real-world images as the real testset. Concretely, we first search more than 1,500 candidate images with real weather noise from several commonly used search engines like Google, Baidu and Bing. Then, candidate images are further filtered according to the image content, image quality and noise types, resulting in more than 600 test images with balanced weather types and diverse scenes. For labeling the ground truth saliency masks, seven professional annotators are organized to independently mark salient objects. Based on the decision records of all annotators, we specify the common object voted by at least half of the annotators as the final salient object. Next, we carefully generate pixel-wise binary masks for these selected objects using Photoshop. Finally, 554 real images with the agreed salient objects and corresponding high-quality pixel-wise masks are obtained. 

\subsection{Dataset Statistics}
To provide in-depth insights into the WXSOD dataset~(\textit{i.e.}, synthesized train set, synthesized test set, and real test set), comprehensive statistics are performed to show its characteristics of diversity and complexity.

\textbf{1) Number and size of salient objects.}
The WXSOD dataset contains different numbers and sizes of salient objects. First, we count the number of images with 1, 2, $\geq$3 salient objects, as shown in Table~\ref{Statistical analysis of size and number}. For all three sets, the vast majority of scenarios~(90.7\%, 95.6\%, 97.2\%) have only one salient object. In addition, we define the size of the salient object, \textit{i.e.}, the proportion of the pixels of the salient object in the mask, and then divide the salient object size into three categories: large, middle, and small. As shown in Table~\ref{Statistical analysis of size and number}, major salient objects tend to be of small and middle size.

\begin{table}[!h]
\centering
\caption{Statistical analysis of object size and numbers in WXSOD. Large:(object size $\geq$30\%), Middle:(5\%$\textless$object size $\textless$30\%), Small:(object size $\leq$5\%).}
\label{Statistical analysis of size and number}
\resizebox{0.65\linewidth}{!}{
\begin{tabular}{ccccccccc}
\hline
\multirow{2}{*}{Sets} &  & \multicolumn{3}{c}{Object Size} &  & \multicolumn{3}{c}{Object Number} \\ \cline{3-5} \cline{7-9} 
&  &Small   & Middle    & Large    &  & 1         & 2         & $\geq$3         \\ \hline
Training      &  &      3825    &    7974       &   1092       &&    9286       &    2128       &   1477        \\
Synthesized       &  &     420     &    976       &    104    &&     1106      &      263     &    131       \\
Real              &  &   223       &      300     &     31     &&       388    &    127       &      39     \\ \hline
\end{tabular}}
\end{table}

\textbf{2) Distribution of weather noise.}
We present the quantitative distribution of weather picture categories in WXSOD, as illustrated in Table.~\ref{Statistical analysis images numbers in WXSOD}. Concretely, WXSOD consists of 12,891 synthesized training images, complemented by 1,500 synthesized testing images and 554 real-world testing images. Among them, the number of images for each weather noise category is balanced.

\begin{table}[!h]
\centering
\caption{Statistical analysis of the number of images corresponding to each weather category in WXSOD.}
\label{Statistical analysis images numbers in WXSOD}
\resizebox{0.80\linewidth}{!}{
\begin{tabular}{cccccccccc}
\hline
\multirow{2}{*}{Sets} & \multicolumn{9}{c}{Image Counts}    \\
\cline{2-10}
& Clean & Dark & Light & Rain & Snow & Fog  & \begin{tabular}[c]{@{}c@{}}Rain\\ \&Snow\end{tabular} & \begin{tabular}[c]{@{}c@{}}Rain\\ \&Fog\end{tabular} & \begin{tabular}[c]{@{}c@{}}Snow\\ \&Fog\end{tabular} \\\hline
Training           & 631   & 1535 & 1531  & 1524 & 1547 & 1534 & 1494    & 1562    & 1533    \\
Synthesized        & 167   & 156  & 166   & 179  & 172  & 166  & 166     & 172     & 156   \\
Real               & --    & 125  & 93    & 120  & 90   & 126  & --      & --   & --    
\\ \hline
\end{tabular}
}
\end{table}

\begin{figure}[!h]
\centering
\includegraphics[width=0.65\linewidth]{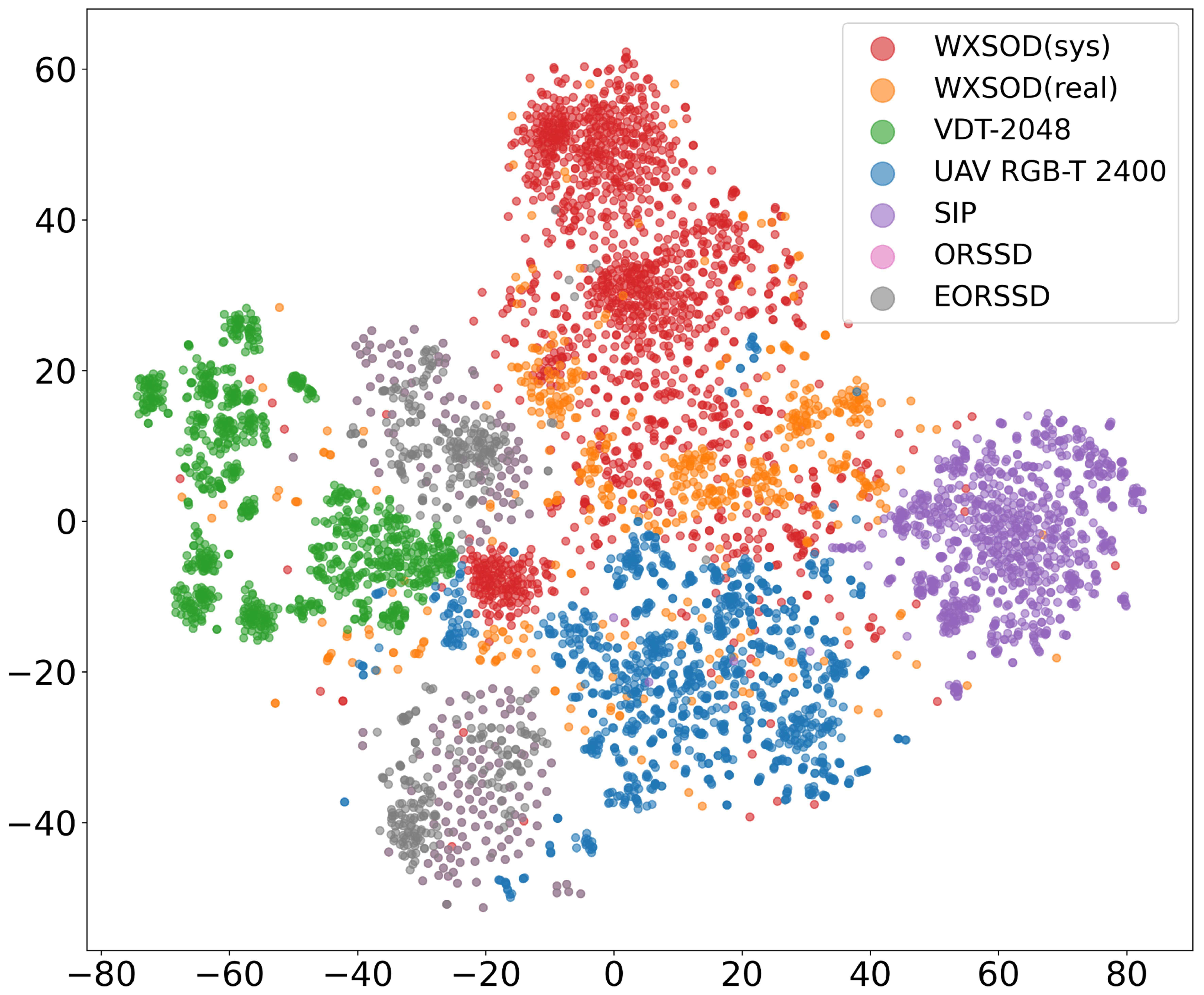}
\caption{Visualization of various datasets using t-SNE.}
\label{t-sne}
\end{figure}  

\subsection{Comparison with Existing Datasets}
To clarify domain differences among various datasets, we visualize the distributions of 7 test sets (involving 6 datasets, with WXSOD including synthetic and real subsets) via the t-SNE, as shown in Fig.~\ref{t-sne}. Specifically, ORSSD, EORSSD, VDT-2048, and SIP datasets, containing only relatively clean data, are limited to local regions in the feature space. The UAV RGB-T 2400 dataset introduces partial noise types (\textit{e.g}, overexposure and dark), and its distribution is adjacent to that of WXSOD. However, WXSOD exhibits wider coverage (especially the real test set). This indicates that after introducing weather noise, WXSOD can provide more comprehensive and challenging data for evaluating models in weather-noise scenarios. Additionally, the partial coupling of the synthetic and real test sets of WXSOD in the feature space, also validates the feasibility of using synthetic noise to expand the data volume.

\section{Methods}\label{sec:Methods}
In this section, we detail the proposed WFANet. In Section~\ref{sec:Network Overview}, we give an overview of the proposed WFANet. Section~\ref{sec:Weather Prediction Branch} details the weather prediction branch. After that, we detail the saliency detection branch presented in Section~\ref{sec:Saliency Detection Branch}.

\begin{figure}[!h]
\centering
\includegraphics[width=1.0\linewidth]{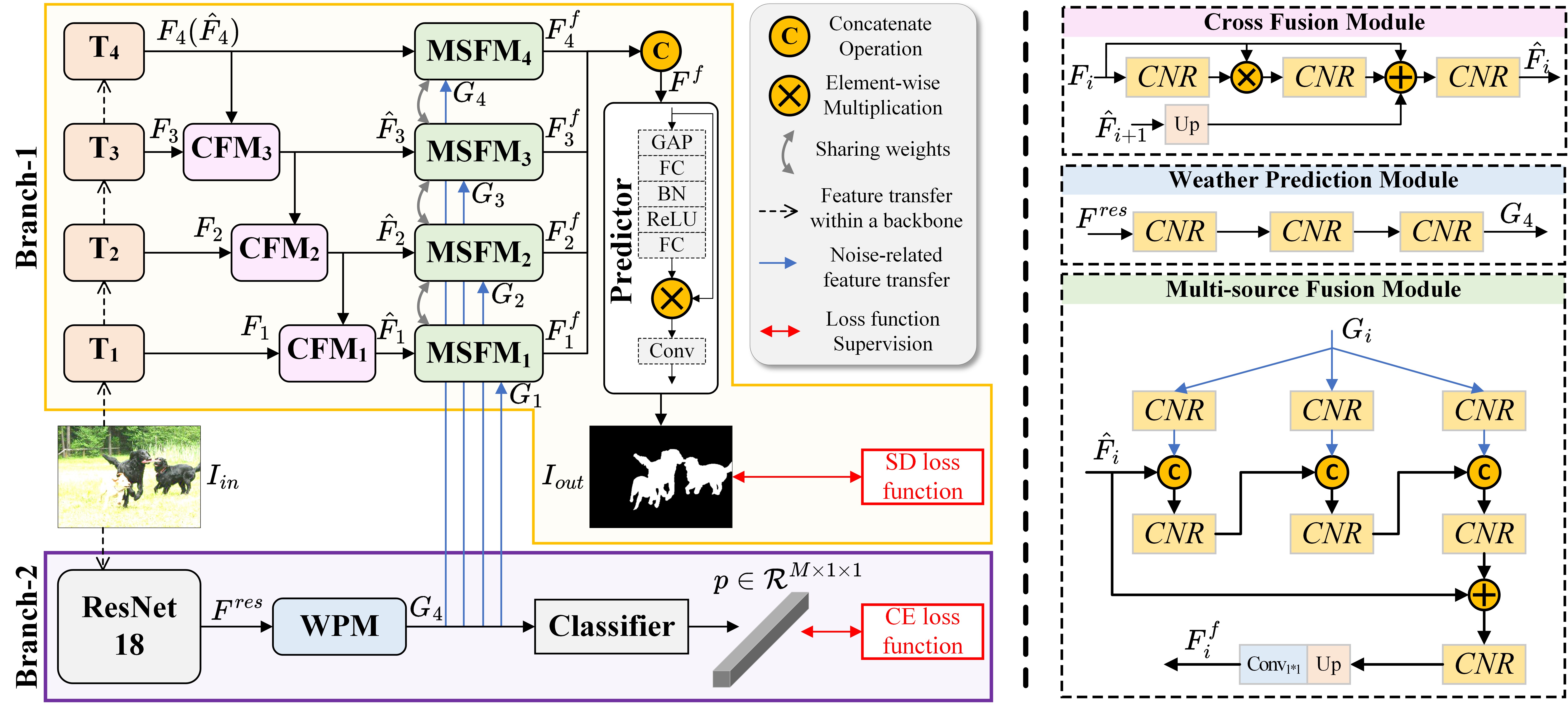}
\caption{The overall architecture of the proposed WFANet, comprising two branches: a weather prediction branch dedicated to learning noise-related features, and a saliency detection branch for SOD. 
}
\label{framework}
\end{figure}  

\subsection{Network Overview}\label{sec:Network Overview}
In Fig.~\ref{framework}, we illustrate the overall framework of WFANet, consisting of a weather prediction branch and a saliency detection branch.
Specifically, the input image $I_{in}\in\mathcal{R}^{3\times H\times W}$ is first fed into the weather prediction branch~(Branch-2) to learn noise-related feature representations. We employ the lightweight ResNet18~\cite{he2016deep} to map $I_{in}$ into high-level features $F^{res}$, followed by a weather prediction module~(WPM) that encodes $F^{res}$ into noise-related embeddings $G_{4}$. To enforce correlation between $G_{4}$ and the noise category of $I_{in}$, a classifier is introduced to predict the weather noise category of $I_{in}$ from $G_{4}$. Notably, $G_{4}$ is sequentially upsampled to form a feature set $\{G_{i}\}_{i=1}^{4}$, which serves as auxiliary input to the saliency detection branch.
Meanwhile, $I_{in}$ is fed into the saliency detection branch~(Branch-1), where it is processed by the backbone PVTv2-b~\cite{wang2021pyramid} to generate multi-scale semantic features $\{F_i\}_{i=1}^4$. Subsequently, adjacent semantic features $F_{i}$ and $F_{i+1}$ are fed into a cross-fusion module~(CFM) to produce aggregated features $\hat{F}_i$, with $F_4=\hat{F}_4$.
Then, we design multi-source fusion module~(MSFM) to adaptively fuse semantic features $\hat{F}_{i}$ and noise-related features $G_{i}$, yielding fused features $F_{i}^{f}$. Finally, the fused features from four MSFMs are concatenated into $F^{f}$, which is used to generate the final high-quality saliency map $I_{out}$ via a predictor.

\subsection{Weather Prediction Branch}\label{sec:Weather Prediction Branch}
As mentioned above, diverse weather-induced degradations (\textit{e.g.}, rain streaks, haze, snow particles) heterogeneously compromise image integrity, significantly impeding saliency detection accuracy. Existing single-branch methods fail to explicitly model these degradation-specific features. We argue that learning noise-related feature representations is beneficial for enhancing the precision of salient object detection. Thus, WFANet incorporates a weather prediction branch that extracts noise-related features via a multi-class classification task. Specifically, given an input image $I_{in}$, we first utilize ResNet18 to extract the corresponding high-level feature $F^{res}$:
\begin{equation}
F^{res}=\mathcal{T}_{extractor}^{resnet}(I_{in})
\end{equation}
where $\mathcal{T}_{extractor}^{resnet}(\cdot)$ denotes the feature extractor based on ResNet18. Then, we design a weather prediction module~(WPM) to encode the high-level feature $F^{res}$:
\begin{equation}\label{eq2}
G_{4}=\mathcal{T}_{WPM}(F^{res})
\end{equation}
where $\mathcal{T}_{WPM}(\cdot)$ denotes the feature encoding process using WPM, and $G_{4}\in\mathcal{R}^{256\times\frac{H}{32}\times\frac{W}{32}}$ denotes the output feature. To explicitly constrain the correlation between $G_{4}$ and weather noise, $G_{4}$ is fed into a classifier to predict weather types:
\begin{equation}
p=\mathcal{T}_{classifier}(G_{4})
\end{equation}
where $\mathcal{T}_{classifier}(\cdot)$ denotes the classifier module; $p\in\mathcal{R}^{M\times1\times1}$ denotes the prediction vector; $M$ denotes the vector length, which is determined by the total number of weather categories, with $M=9$.

We emphasize that the weather category prediction process is supervised, with cross-entropy loss employed to optimize parameters of Branch-2:
\begin{equation}
    \mathcal{L}_{CE}=-\sum_{m=1}^{M}y_{m}\cdot log(\hat{p}_{m})
\end{equation}
where $y_{m}$ denotes the one-hot encoding of the sample, with $y_{m}=1$ if the sample belongs to the $m$-th class and $y_{m}=0$ otherwise; $\hat{p}_{m}$ denotes the predicted probability for the $m$-th class after softmax normalization:
\begin{equation}
    \hat{p}_{m}=\frac{e^{p_{m}}}{\sum_{i=1}^{M}e^{p_{i}}}
\end{equation}
where $p_{m}$ denotes the logit output for the $m$-th class.

Next, we present the details of the WPM and classifier. As shown in Fig.~\ref{framework}, WPM is formed by three stacked basic units $CNR$ which contains a $3\times3$ convolutional layer~($Conv$), a batch normalization layer~($BN$) and a activation layer~($ReLU$). The entire process is formulated as follows:
\begin{equation}
    CNR(x)=ReLU(BN(Conv_{3\times 3}(x)))
\end{equation}
Therefore, Eq~(\ref{eq2}) can be rewritten as:
\begin{equation}
    G_{4}=\mathcal{T}_{WPM}(F^{res})=CNR_{3}(CNR_{2}(CNR_{1}(F^{res})))
\end{equation}

Benefiting from the guidance of the multi-classification task, $G_{4}$ can be regarded as a noise-related feature, which is further fed into the Branch-1 as supplementary input.
To maintain feature resolution consistency across branches, $G_{4}$ is upsampled via bi-cubic interpolation to generate the feature set $\{G_{i}\}_{i=1}^{4}$:
\begin{equation}
    G_{i}=\mathcal{U}_{\times2^{4-i}}(G_{4}), \quad i\in{1,2,3}
\end{equation}
where the height and width of $G_{i}$ are $\frac{H\cdot2^{4-i}}{32}$ and $\frac{W\cdot2^{4-i}}{32}$. $\mathcal{U}_{\times}(\cdot)$ denotes the upsampling operation, where the subscript indicates the upsampling factor.

In addition, the classifier consists of following layers: a fully connected layer~($FC$), a batch normalization layer~($BN$), a dropout layer~($Dropout$), and a classification layer~($Cls$), which is a fully-connected layer. The classifier is deployed to predict the weather-tag of each image. Therefore, the number of output neurons of $Cls$ is $M$, which is equal to the number of weather categories.

\subsection{Saliency Detection Branch}\label{sec:Saliency Detection Branch}
The remainder of WFANet is defined as saliency detection branch, which includes three parts: cross fusion module, multi-source fusion module and saliency predictor.

First, we employ PVTv2-b to extract the multi-scale semantic feature $\{F_{i}\}_{i=1}^{4}$ from the input image $I_{in}$, where each $F_{i}$ has spatial resolution $\frac{H\cdot2^{4-i}}{32}\times\frac{W\cdot2^{4-i}}{32}$. To establish spatial dependencies across scales, we design a cross fusion module that fuses adjacent semantic features:
\begin{equation}
\hat{F}_{i}=\mathcal{T}_{CFM}(F_{i},\hat{F}_{i+1}),\quad i\in \{3,2,1\}
\end{equation}
with the fusion operation $\mathcal{T}_{CFM}(\cdot,\cdot)$ defined as:
\begin{equation}
\hat{F}_i=CNR_{3}\left(\mathcal{U}_{\times2}\left(\hat{F}_{i+1}\right)+CNR_{2}\left(CNR_{1}\left(F_i\right) \odot F_i\right)+F_i\right)
\end{equation}
where $\odot$ denotes pixel-wise multiplication. Inspired by~\cite{yi2024gponet}, we employ feature gating via element-wise multiplication between outputs and input features to suppress information redundancy.
Note that the highest-scale feature $F_{4}$ is identical to $\hat{F}_{4}$.

Next, the multi-source fusion module is designed to integrate semantic and noise-related features from CFM and WPM, respectively.
Four MSFMs with shared weights are employed to process features at each scale in parallel. For a single MSFM, a progressive fusion strategy is adopted to integrate the two input features.
As depicted in Fig.~\ref{framework}, with $\hat{F}_{i}$ and $G_{i}$ as inputs, the MSFM processing can be defined as:
\begin{equation}
\left\{\begin{array}{l}
\hat{F}_{i}^{1} = CNR(\copyright[CNR(G_{i}), \hat{F}_{i}]) \\
\hat{F}_{i}^{2} = CNR(\copyright[CNR(G_{i}), \hat{F}_{i}^{1}]) \\
\hat{F}_{i}^{3} = CNR(\copyright[CNR(G_{i}), \hat{F}_{i}^{2}]) \\
{F}_{i}^{f} = Conv_{1\times1}(\mathcal{U}_{in}(CNR(\hat{F}_{i}^{3}+\hat{F}_{i}))) \\
\end{array}\right.
\end{equation}
where $\mathcal{U}_{in}(\cdot)$ denotes the operation that upsamples the feature to the same resolution as the input image $I_{in}$; \copyright$[\cdot,\cdot]$ denotes the feature concatenation operation;
$\{\hat{F}_{i}^{1},\hat{F}_{i}^{2},\hat{F}_{i}^{3}\}$ represents intermediate features.
All fused feature ${F}_{i}^{f}\in\mathcal{R}^{1\times H\times W}$ are concatenated along the channel dimension  to form the final feature map $F^{f}\in\mathcal{R}^{4\times H\times W}$ for saliency prediction. The process can be written as:
\begin{equation}
F^{f}=\copyright[F_{1}^{f},F_{2}^{f},F_{3}^{f},F_{4}^{f}]
\end{equation}

Finally, we design a predictor that takes the feature $F^f$ as input to predict salient objects. The predictor employs a channel attention mechanism to adaptively fuse multi-scale information within $F^f$, followed by a $3\times3$ convolutional layer and sigmoid activation to generate the single-channel saliency map $I_{out}\in\mathcal{R}^{1\times H\times C}$:
\begin{equation}
    I_{out} = \mathcal{T}_{pre}(F^{f})
\end{equation}
where $\mathcal{T}_{pre}(\cdot)$ denotes the operation of the predictor. 
We adopt the hybrid loss $\mathcal{L}_{Mix}$~(\textit{i.e.}, BCE loss, SSIM loss and IoU loss)~\cite{qin2019basnet} to supervise Branch-1.

As shown in Fig.~\ref{framework}, the predictor comprises a global average pooling layer~($GAP$), two fully connected layers~($FC$), a batch normalization layer~($BN$) and a ReLU activation layer. The entire process can be written as follows:
\begin{equation}
\mathcal{T}_{pre}(F^{f})= \sigma(Conv(FC_{2}(BN(FC_{1}(GAP(F^{f}))))\otimes F^{f}))
\end{equation}
where $\sigma(\cdot)$ means sigmoid function, and the numbers of output neurons for $FC_1$ and $FC_2$ are 256 and 4, respectively.

\section{Experiments}\label{sec:Experiments}

\subsection{Implementation Details}\label{Implementation Details}
We employ PVTv2-b and ResNet18 as the backbone networks for Branch-1 and Branch-2, respectively. The model is optimized using Adam with initial learning rate \text{$1e^{-4}$} (halved every 30 epochs), batchsize 6 and input resolution 384$\times$384. Training converged within 50 epochs with data augmentation including random flipping, rotation and boundary clipping. In the testing phase, all quantitative results are measured using original images, including both synthesized and real test set.

\subsection{Evaluation Metrics}\label{Evaluation Metrics}
To fairly evaluate all compared models, we evaluate performances of comparison methods on 10 most extensive evaluation metrics, including S-measure~($S$)~\cite{fan2017structure}, mean absolute error~($MAE$)~\cite{perazzi2012saliency}, E-measure~($E_{\xi}^{adb}$, $E_{\xi}^{mean}$, $E_{\xi}^{max}$)~\cite{fan2018enhanced}, F-measure~($F_{\beta}^{adb}$, $F_{\beta}^{mean}$, $F_{\beta}^{max}$)~\cite{achanta2009frequency}, F-measure curve and Precision-Recall~(PR curve)~\cite{achanta2009frequency}.

To compare computational costs of different models, we report the number of learnable parameters~($Params$), multiply–accumulate operations~($MACs$), and frames per second ($FPS$). Note that the $MACs$ and $FPS$ are measured on a 384$\times$384 image, using an NVIDIA 3090 GPU.

\subsection{Comparison with the State-of-the-art Methods}\label{Comparison with the State-of-the-art Methods}
We include 18 deep learning-based state-of-the-art methods in our benchmark for advanced evaluations, which consists of 4 backbone-free methods~(MINet~\cite{shen2024minet}, FSMINet~\cite{shen2022fully}, ADMNet~\cite{zhou2024admnet} and 
TRACER~\cite{lee2022tracer}), and the remaining 14 backbone-based methods~(SEANet~\cite{li2023lightweight}, CorrNet~\cite{gongyangli2022lightweight}, A3Net~\cite{cui2023autocorrelation}, HDNet~\cite{lu2024low}, AESINet~\cite{zeng2023adaptive}, ICONet~\cite{zhuge2022salient}, DCNet~\cite{zhu2025dc}, MEANet~\cite{liang2024meanet}, SAFINet~\cite{luo2024spatial}, MSRMNet~\cite{liu2024msrmnet}, TCGNet~\cite{liu2023tcgnet}, GeleNet~\cite{li2023salient}, GPONet~\cite{yi2024gponet} and our WFANet).
For a fair comparison, the prediction results of all models are generated by running the source codes with their default parameter settings and a unified loss function.

\begin{table}[!h]
\centering
\caption{Quantitative comparison results of $S$, $MAE$, $E^{adb}$, $E^{mean}$, $E^{max}$, $F_{\beta}^{adb}$, $F_{\beta}^{mean}$ and $F_{\beta}^{max}$ on the synthesized test set. The computational costs include the parameters~($Params$), $MACs$ and $FPS$ for each method. Here, “$\uparrow$” (“$\downarrow$”) means that the larger (smaller) the better. The best three results in each row are marked in red, green, and blue, respectively.}
\label{Quantitative result on synthesized data}
\resizebox{1.0\linewidth}{!}{
\begin{tabular}{ccccccccccccccc}
\hline
\multirow{2}{*}{Methods} & \multirow{2}{*}{Backbone} &  & \multicolumn{8}{c}{Quantitative results} &  & \multicolumn{3}{c}{Computational costs} \\ \cline{4-11} \cline{13-15} 
&  &  & $MAE\downarrow$   & $S\uparrow$  & 
$F_{\beta}^{adb}\uparrow$  & $F_{\beta}^{mean}\uparrow$  & $F_{\beta}^{max}\uparrow$  & 
$E_{}^{adb}\uparrow$  & $E_{}^{mean}\uparrow$  & $E_{}^{max}\uparrow$  & 
& $Params$(M)   & $MACs$(G) & $FPS$   \\ \hline
MINet\cite{shen2024minet}  & --  &  &  0.0775   &  0.7403    & 0.6207   &  0.6219  &  0.6501  &   0.8130 &  0.7976  & 0.8263   &  &   \First{0.36}            &      \First{0.27}   &  \Third{377.37}  \\
FSMINet\cite{shen2022fully}   & --  &  &  0.0473   &  0.8306    & 0.7558   &  0.7635  &  0.7805  &  0.8781  &  0.8693  & 0.8808   &  &   3.56            &    11.82    & 93.32 \\
ADMNet\cite{zhou2024admnet} &  --  &  & 0.0770    &   0.7497   &  0.6349  &  0.6349  &  0.6594  &  0.8213  &  0.8061  &  0.8307  &  &   \Second{0.94}    &      \Second{0.83}     &      \First{653.78}    \\
TRACER\cite{lee2022tracer}   & EfficientNet-b0  &  &  0.0280   &   0.8898   &   0.8285 &  0.8473  &   0.8671 &  0.9310  &  \Third{0.9319}  &  0.9414  &  &  3.89    &      \Third{3.12}    & \Second{444.92}  \\
SEANet\cite{li2023lightweight} &  MobileNet-V2 &  &  0.0332   &   0.8694   &  0.8129  &  0.8210  &  0.8391  &   0.9201 &  0.9136  &  0.9240  &  &   \Third{2.74}            &   3.21    & 309.14  \\
CorrNet\cite{gongyangli2022lightweight} & VGG16  &  &   0.0613  &   0.7652   &  0.7378  &  0.6911  &  0.7270  &   0.8403 &  0.7729  &  0.8281  &  &   4.06            &   47.95  &  60.14  \\
AESINet\cite{zeng2023adaptive}  &  VGG16  &  &  0.0337   &  0.8618    &  0.8075  & 0.8141   &  0.8247  &  0.9096  & 0.9039   & 0.9085   &  &    41.04           &  155.12   &  45.78  \\
A3Net\cite{cui2023autocorrelation} &  ResNet50 &  &  0.0327   &   0.8784   & 0.8158   &  0.8286  &   0.8472 &  0.9176  &  0.9131  & 0.9253   &  &   22.70            &    27.94     &  228.08      \\
HDNet\cite{lu2024low} &  ResNet50    &  &  0.0397   &   0.8537   & 0.7875   &  0.7964  &   0.8162 &  0.8990  &  0.8901  & 0.9053   &  &   24.67     &     26.19       &    196.14   \\
ICONet\cite{zhuge2022salient}  &  ResNet50   &  &  0.0433   &   0.8411   &  0.7496  &  0.7704  &  0.7957  &   0.8924 &  0.8925  & 0.9078   &  &   33.03            &     24.95     &   270.19     \\
DCNet\cite{zhu2025dc}    &     ResNet34        &  &   0.0320  &  0.8792    &  0.8336  &  0.8372  & 0.8498   &  0.9179  &  0.9086  &  0.9221  &  &    88.95           &      120.29  &    94.27         \\
MEANet\cite{liang2024meanet}  &    MobileNet-V2            &  &  0.0332   &    0.8772  &  0.8174  &  0.8294  &  0.8502  &  0.9225  &  0.9185  &  0.9288  &  &   3.27            &    13.48  &       235.27    \\
SAFINet\cite{luo2024spatial}  &    MobileNet-V2          &  &  0.0384   &    0.8643  &  0.8049  &  0.8105  &  0.8303  &  0.9073  &  0.8982  &  0.9120  &  &    3.12     &   13.67  &  232.60  \\
MSRNet\cite{liu2024msrmnet} &     Swin-B         &   &  0.0275   &  0.8895    &  0.8361  &  0.8493  &  0.8657  &  0.9267  &  0.9265  &  0.9346  &  &       89.66     &   54.30       &   85.12  \\
TCGNet\cite{liu2023tcgnet}  &    HRNet           &  &  0.0317   &  0.8832    &  0.8278  &  0.8388  &  0.8576  &   0.9228 &  0.9180  & 0.9288   &  &   70.26      &    60.14    &     52.84       \\
GeleNet\cite{li2023salient} &   PVTv2-b             &  &  \Second{0.0239}   &   \Second{0.9038}   &  \Second{0.8538}  &  \Second{0.8675}  &  \Second{0.8868}  &  \Second{0.9418}  &  \Second{0.9410}  &  \Second{0.9499}  &  &  25.45    &      13.91  &   250.54         \\
GPONet\cite{yi2024gponet}  &      PVTv2-b        &  &   \Third{0.0266}  &   \Third{0.8962}   &  \Third{0.8445}  &  \Third{0.8568}  &  \Third{0.8774}  &   \Third{0.9352} &  0.9318  &   \Third{0.9425}  &  &     24.86          &      11.52       &    83.49    \\
Ours   &    \begin{tabular}[c]{@{}c@{}}PVTv2-b \\ \& ResNet-18\end{tabular}   &  &   \First{0.0229}  &  \First{0.9051}    &   \First{0.8601}  &  \First{0.8713}  &  \First{0.8888}  &  \First{0.9464}  &  \First{0.9443}  &  \First{0.9523}  &  &    50.87           &    112.63         &    71.64    \\ \hline
\end{tabular}
}
\end{table}

\textbf{1) Quantitative Comparison.}
To comprehensively evaluate our WFANet on WXSOD dataset, Tables~\ref{Quantitative result on synthesized data} and \ref{Quantitative result on real data} present quantitative results~(including $MAE$, $S$, $F_{\beta}^{adb}$, $F_{\beta}^{mean}$, $F_{\beta}^{max}$, $E_{\xi}^{adb}$, $E_{\xi}^{mean}$ and $E_{\xi}^{max}$ ) and computational costs~($Params$ and $MACs$).

On the synthesized test set~(Table~\ref{Quantitative result on synthesized data}), WFANet achieves the lowest $MAE$ of 0.0229 and the highest $S$-measure of 0.9051, indicating excellent pixel-wise accuracy and structural similarity with GT. WFANet also achieves top-ranked performance across other six metrics, demonstrating its superior performance.
For instance, compared to GPONet with same backbone network, our method reduces $MAE$ by 13.9\%.

On the real test set (Table~\ref{Quantitative result on real data}), WFANet maintains its leading position with the lowest $MAE$ of 0.0159 and the highest $S$-measure of 0.9248.  It also achieves top scores in $F_{\beta}$ and $E$ metrics, further validating its robustness and effectiveness. Against GeleNet and GPONet, our method reduces $MAE$ by 17.1\% and 15.4\%, highlighting better robustness to real-world noise. For backbone-free methods~(MINet, FSMINet, etc.), our model outperforms them across all metrics. 
While WFANet's dual-branch architecture incurs higher $Params$ and $MACs$, its optimal detection accuracy validates this design as a judicious performance-efficiency trade-off. 
Future study could focus on optimizing the computational efficiency while preserving competitive performance.

\begin{table}[!h]
\centering
\caption{Quantitative comparison results on the real test set, and the computational costs of each method.}
\label{Quantitative result on real data}
\resizebox{1.0\linewidth}{!}{
\begin{tabular}{ccccccccccccccc}
\hline
\multirow{2}{*}{Methods} & \multirow{2}{*}{Backbone} &  & \multicolumn{8}{c}{Quantitative results} &  & \multicolumn{3}{c}{Computational costs} \\ \cline{4-11} \cline{13-15} 
& &  & $MAE\downarrow$   & $S\uparrow$  & 
$F_{\beta}^{adb}\uparrow$  & $F_{\beta}^{mean}\uparrow$  & $F_{\beta}^{max}\uparrow$  & 
$E_{}^{adb}\uparrow$  & $E_{}^{mean}\uparrow$  & $E_{}^{max}\uparrow$  & 
& $Params$(M)     & $MACs$(G)  & $FPS$ \\ \hline
MINet\cite{shen2024minet}  & --  &  &   0.0549  &   0.7710   &  0.6265  &  0.6425  &  0.6748  &   0.8278 &  0.8205  &  0.8442  &  &  \First{0.36}  &        \First{0.27} &  \Third{377.37}  \\
FSMINet\cite{shen2022fully} & --  &  &  0.0264   &   0.8869   &  0.8123  &  0.8284  &  0.8541  & 0.9239   &  0.9231  &  0.9353  &  &   3.56   &    11.82    &  93.32 \\
ADMNet\cite{zhou2024admnet} & --  &  &  0.0544   &   0.7839   &  0.6676  &  0.6683  & 0.6921   &   0.8503 &  0.8262  &  0.8554  &  &   \Second{0.94}      &    \Second{0.83}  & \First{653.78} \\
TRACER\cite{lee2022tracer}&   EfficientNet-b0    &  &  0.0227   &  0.8961    &  0.8225  &  0.8390  &  0.8586  &  0.9230  &  0.9224  & 0.9322   &  &    3.89           &    \Third{3.12}     &  \Second{444.92} \\
SEANet\cite{li2023lightweight}&   MobileNet-V2  &  &   0.0239  &  0.8978    &  0.8192  &  0.8390  &  0.8706  &   0.9261 & 0.9290   &  0.9408  &  &  \Third{2.74}      &   3.21          &   309.14    \\
CorrNet\cite{gongyangli2022lightweight} &  VGG16   &  &  0.0534   &   0.7674   &  0.7288  &  0.6629  &  0.7223  & 0.8363   &  0.7488  & 0.8268   &  &  4.06             &     47.95        &   60.14     \\
AESINet\cite{zeng2023adaptive} &  VGG16  &  &  0.0251   &   0.8845   &  0.8202  &  0.8359  &  0.8602  &  0.9267  &  0.9304  &  0.9367  &  &   41.04            &    155.12  & 45.78  \\
A3Net\cite{cui2023autocorrelation} & ResNet50  &  &   0.0263  &   0.8909   &  0.8172  &  0.8350  &  0.8586  &   0.9207 &  0.9199  & 0.9329   &  &   22.70            &    27.94   &  228.08  \\
HDNet\cite{lu2024low} &  ResNet50   &  &   0.0266  &  0.8938    & 0.8179   &  0.8447  &  0.8734  & 0.9163   & 0.9219   &  0.9361  &  &  24.67             &     26.19         &    196.14   \\
ICONet\cite{zhuge2022salient} &  ResNet50  &  &   0.0327  &   0.8619   &  0.7595  &  0.7885  &  0.8209  &  0.9007  &  0.9117  & 0.9276   &  &   33.03            &      24.95   &  270.19 \\
DCNet\cite{zhu2025dc} & ResNet34   &  &  0.0215   &  0.9058    &  \Third{0.8564}  &  0.8672  &  0.8847  &  0.9425  &  0.9357  &   0.9456 &  &   88.95            &   120.29    &     94.27   \\
MEANet\cite{liang2024meanet}  &    MobileNet-V2     &  &  0.0236   &   0.9063   & 0.8284   &  0.8472  &  0.8803  &  0.9293  &  0.9286  &  0.8418  &  &  3.27             &    13.48   &    235.27   \\
SAFINet\cite{luo2024spatial} &  MobileNet-V2    &  &  0.0217   &    0.9050  &  0.8274  &  0.8438  &  0.8723  &  0.9265  &  0.9242  &  0.9407  &  &   3.12            &    13.67       &   232.60  \\
MSRNet\cite{liu2024msrmnet} &  Swin-B   &  &   0.0198  &   0.9091   &   \Second{0.8595} &  \Third{0.8716}  &  0.8881  &  \Second{0.9453}  &  \Third{0.9432}  &  0.9516  &  &   89.66    &   54.30   & 85.12  \\
TCGNet\cite{liu2023tcgnet} &  HRNet    &  &   \Second{0.0185}  &   \Third{0.9185}   &  0.8529  &  \Second{0.8739}  &  \Second{0.8992}  &   \Third{0.9429} &  \Second{0.9476}  & \Second{0.9594}   &  &   70.26            &   60.14         &    52.84     \\
GeleNet\cite{li2023salient}&   PVTv2-b    &  &   0.0192  &  \Second{0.9186}    &  0.8442  &  0.8698  &  \Third{0.8932}  & 0.9316   &  0.9425  &  \Third{0.9528}  &  &     25.45   &    13.91     &   250.54     \\
GPONet\cite{yi2024gponet} &  PVTv2-b   &  &  \Third{0.0188}   &   0.9098   &  0.8434  &  0.8604  &  0.8830  &   0.9362 &  0.9364  &  0.9468  &  &   24.86            &      11.52 &   83.49   \\
Ours &    \begin{tabular}[c]{@{}c@{}}PVTv2-b \\ \& ResNet-18\end{tabular}  &  &  \First{0.0159}   &   \First{0.9248}   &  \First{0.8617}  &  \First{0.8828}  &  \First{0.9080}  & \First{0.9494}   &  \First{0.9529}  &  \First{0.9617}  &  &      50.87     &     112.63     &  71.64   \\ \hline
\end{tabular}
}
\end{table}

\begin{figure}[!h]
\centering
\includegraphics[width=1.0\linewidth]{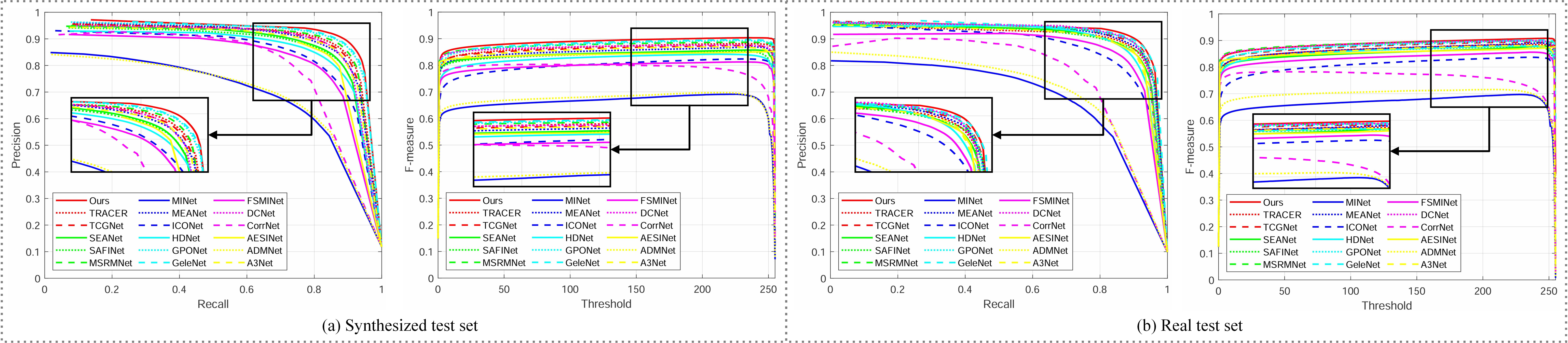}
\caption{Quantitative evaluation of different saliency models: (a) presents PR curves and F-measure curves on the synthesized test set, while (b) presents PR curves and F-measure curves on the real test set.}
\label{Curves}
\end{figure}  

For comprehensive quantitative evaluation, Fig.~\ref{Curves} presents Precision-Recall~(PR) and F-measure curves comparing WFANet against other 17 models. It can be observed that the PR curve of WFANet is the closest one to the upper right corner, and the area below the F-measure curve of WFANet is also the largest one. These experiments results prove the effectiveness and superiority of WFANet.

\begin{figure}[!t]
\centering
\includegraphics[width=1.0\linewidth]{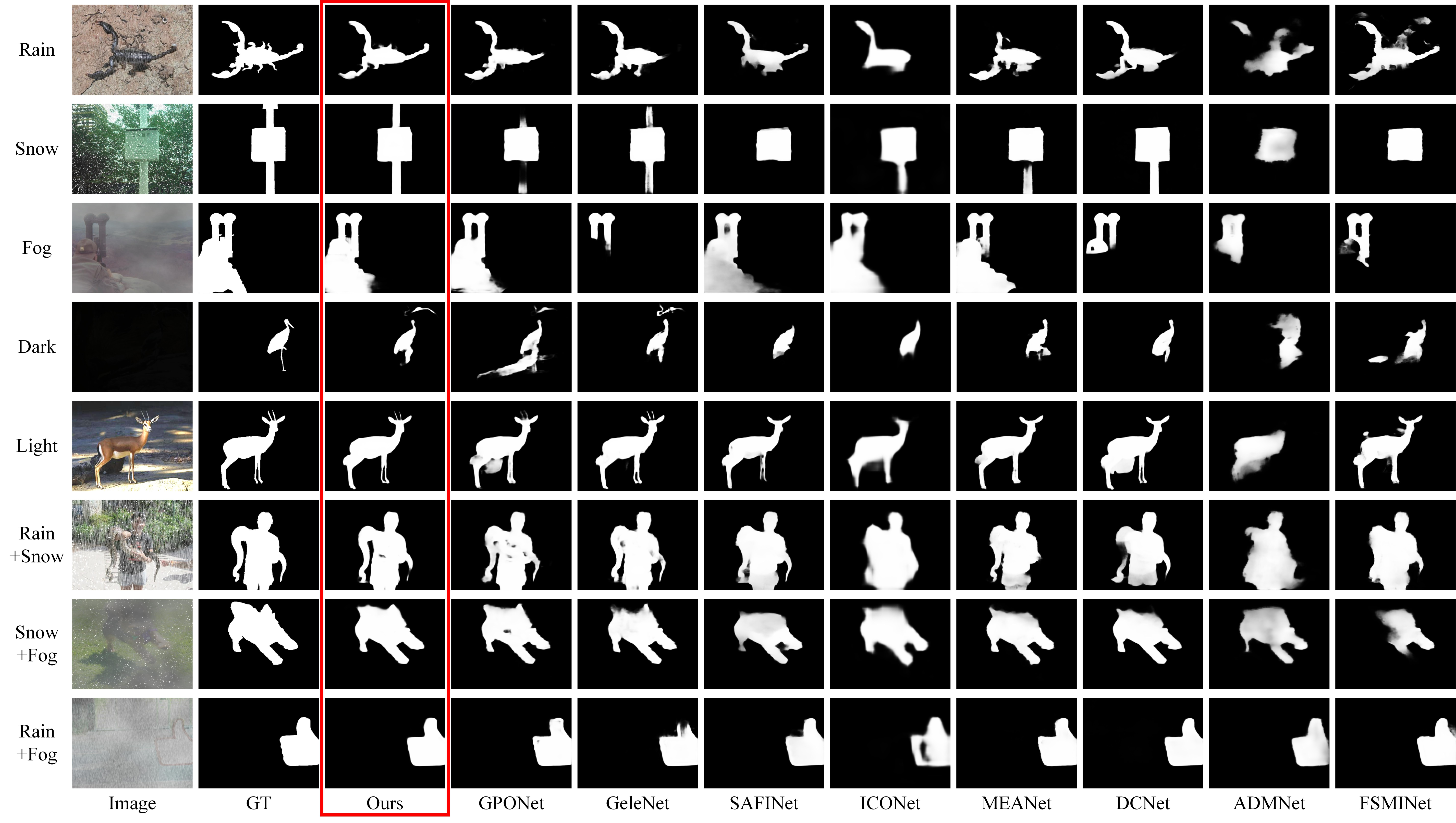}
\caption{Visual comparisons of different methods on the synthesized test set.}
\label{Qualitative result of synthesized noise}
\end{figure}
\begin{figure}[!t]
\centering
\includegraphics[width=1.0\linewidth]{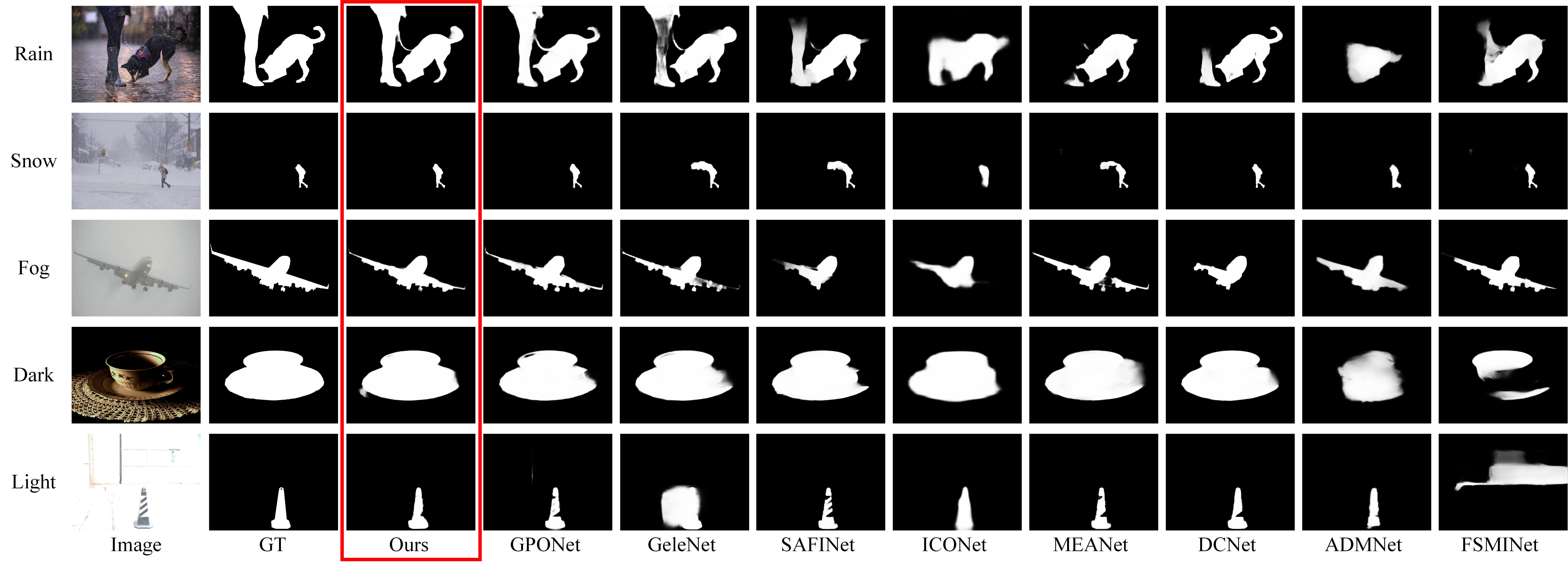}
\caption{Visual comparisons of different methods on the real test set.}
\label{Qualitative result of real noise}
\end{figure}

\textbf{2) Qualitative Comparison.}
To qualitatively make a comparison for saliency models, Fig.~\ref{Qualitative result of synthesized noise} and Fig.~\ref{Qualitative result of real noise} provide visual comparisons for both synthesized and real samples.
Visualizations reveal that comparative models fail to capture complete contours of salient objects or misclassify background regions as targets under noise interference. In contrast, WFANet generates saliency maps closest to GT, robustly preserving target integrity despite weather-induced noise. Notably, WFANet outperforms PVT-backbone counterparts GeleNet and GPONet.
As shown in Columns 4 and 5 of Fig.~\ref{Qualitative result of synthesized noise} and.~\ref{Qualitative result of real noise}, while GeleNet and GPONet can localize salient targets, their segmentation compromises target integrity and introduce irrelevant regions. In contrast, WFANet's dual-branch architecture balances the learning of noise-related features and salient features, enabling adaptive filter meaningful information from extreme weather scenes and retaining precise salient object contours. For instance, in the synthesized snow+fog scene~(Row 7, Fig.~\ref{Qualitative result of synthesized noise}), WFANet segments the entire dog, whereas GPONet misclassifies the inner part of the target as background, and GeleNet confuses noise with target edges, resulting in blurred segmentation boundaries. 
Overall, these results validate the robustness of WFANet in predicting salient targets under adverse weather conditions.

\subsection{Cross Dataset Generalization Validation}\label{Cross Dataset Generalization Validation}

To demonstrate that dataset discrepancies~(\textit{i.e.}, noise types, view perspectives, and image counts) constrain model performance in complex weather scenarios, we retrain WFANet using training sets from various sources, including DUTs, VT5000, VDT2048, and UAV RGB-T 2400. 
To isolate the impact of image counts, we construct a noise-free WXSOD counterpart (denoted WXSOD~(clean)) by removing additional noise.
As shown in Table~\ref{cross verify}, VT5000/VDT2048-trained models perform poorly, due to negligible noise content and limited sample size.
Despite including certain noise~(\textit{e.g.}, low light and overexposure), UAV RGB-T 2400-trained model yield inferior performance, primarily caused by view perspective mismatch with WXSOD.
Notably, WXSOD (clean) and DUTs have comparable sample sizes, but their trained models underperform those trained on WXSOD. This confirms that WXSOD uniquely enhances model performance in complex weather scenarios, enabled by tailored diverse noise.

\begin{table}[!h]
\centering
\caption{Ablation studies of cross dataset validation.}
\label{cross verify}
\resizebox{1.0\linewidth}{!}{
\begin{tabular}{lcccccccccc}
\hline
\multirow{2}{*}{Training set} &  & \multicolumn{4}{c}{Synthesized test set} &  & \multicolumn{4}{c}{Real test set} \\ \cline{3-6} \cline{8-11} 
&  & $MAE\downarrow$ & $S\uparrow$ & $F_{\beta}^{mean}\uparrow$ & $E_{}^{mean}\uparrow$   &  & $MAE\downarrow$ & $S\uparrow$ & $F_{\beta}^{mean}\uparrow$ & $E_{}^{mean}\uparrow$   \\ \hline
DUTs~\cite{wang2017learning}                          &  &       \Third{0.0612}  &  \Second{0.8306} &  \Third{0.7725} &  \Second{0.8651}  &  &   \Second{0.0196}   &  \Second{0.9194} &  \Third{0.8687} &  \Second{0.9420}    \\
VT5000~\cite{tu2022rgbt}                        &  &    0.0676   &  0.7950 &  0.7006 &  0.8342    &  &   0.0506   &  0.8314 &  0.7296 &  0.8644   \\
VDT2048~\cite{song2022novel}                       &  &   0.0844   & 0.6877  & 0.6089  & 0.7480    &  &    0.0574  & 0.7818  & 0.7111  & 0.8333   \\
UAV RGB-T 2400~\cite{song2023modality}                &  &  0.1211  & 0.4960  &  0.1866 & 0.4599    &  &     0.1001   & 0.5506  & 0.3383  &  0.6064  \\
WXSOD (clean)                 &  &      \Second{0.0473}   &  \Third{0.8229} &  \Second{0.7758} &  \Third{0.8567}    &  &        \Third{0.0213}   & \Third{0.9072}  &  \Second{0.8697} & \Third{0.9361}     \\ \hdashline
WXSOD                         &  &      \First{0.0229}  & \First{0.9051}  & \First{0.8713}  &  \First{0.9443}   &  &    \First{0.0159}   &  \First{0.9248} &  \First{0.8828} &  \First{0.9529}   \\ \hline
\end{tabular}
}
\end{table}

\subsection{Ablation Studies}\label{Ablation Studies}

\textbf{1) Model architecture effectiveness.}
To evaluate the effectiveness of the dual-branch architecture in WFANet, we construct three variant models~(designated \#1, \#2 and \#3 Model). The experimental results are presented in Table~\ref{variants result}.


Comparing \#1 Model and \#4 Model reveals that removing the weather prediction branch from WFANet leads to a decline in performance across all quantitative metrics. This indicates the importance of the weather prediction branch in enhancing the WFANet performance. 
The comparison between \#1 Model and \#2 Model demonstrates that incorporating the weather prediction branch, even without its dedicated classification supervision(\textit{i.e.}, operating the branch without explicit noise-class labels), still yields performance improvements over the single-branch baseline. This improvement can be attributed to the fact that, under the unsupervised paradigm, the additional ResNet18 backbone can extract rich visual semantic information, which acts as supplementary cues to optimize the feature representation of the saliency detection branch.
Notably, evolving \#2 Model into the full WFANet (\#4 Model) by reintroducing the classification loss for the weather branch provides a further consistent boost across main evaluation metrics. 
This result confirms the advantage of providing explicit supervised training for the weather branch to specifically capture weather noise-related features.
The comparison between \#3 Model and \#4 Model highlights the importance of the CFM in enhancing the correlation of multi-scale semantic features. This capability effectively boosts the model's detection accuracy.

\begin{table}[!h]
\centering
\caption{Ablation studies of between our WFANet~(\#4) and different variants~(\#1, \#2 and \#3).}
\label{variants result}
\resizebox{1.0\linewidth}{!}{
\begin{tabular}{lcccccccccccccccccccc}
\hline
\multirow{2}{*}{No.} &  & \multicolumn{3}{c}{Different structure} &  & \multicolumn{2}{c}{Different loss} &  & \multicolumn{4}{c}{Synthesized test set} &  & \multicolumn{4}{c}{Real test set} &  & \multicolumn{2}{c}{Computational costs} \\ \cline{3-5} \cline{7-8} \cline{10-13} \cline{15-18} \cline{20-21} 
&  & CFM         & MSFM         & WPM        &  & SD loss          & CE loss         &  & $MAE\downarrow$ & $S\uparrow$ & $F_{\beta}^{mean}\uparrow$ & $E_{}^{mean}\uparrow$    &  & $MAE\downarrow$ & $S\uparrow$ & $F_{\beta}^{mean}\uparrow$ & $E_{}^{mean}\uparrow$   &  & $Parmas$(M)              & $MACs$(G)              \\ \hline
\#1                    &  & $\checkmark$            & $\checkmark$             &            &  & $\checkmark$                 &                 &  &   0.0244   &   0.9022   &   0.8621  &   0.9378  &  &   \Third{0.0171}   &   0.9161   &   0.8673   &   0.9489  &  &       \First{33.39}     &     \First{63.58}   \\
\#2                    &  & $\checkmark$            & $\checkmark$             & $\checkmark$           &  & $\checkmark$                 &                 &  &    \Third{0.0237}  &     \Second{0.9058} &  \Third{0.8640}   &   \Third{0.9422}  &  &    \Second{0.0165}  &    \Third{0.9202}  &   \Third{0.8688}   &  \Second{0.9507}   &  &    50.87      &    112.63   \\
\#3                    &  &             & $\checkmark$             & $\checkmark$           &  & $\checkmark$                 & $\checkmark$                &  &  \First{0.0226}    &  \First{0.9091}    &  \Second{0.8709}   &   \Second{0.9428}  &  &  0.0173    &  \Second{0.9243}    &   \Second{0.8786}   &   0.9499  &  &  \Second{47.33}  &    \Second{98.34}   \\
\#4                    &  & $\checkmark$            & $\checkmark$             & $\checkmark$           &  & $\checkmark$                 & $\checkmark$                &  &    \Second{0.0229}  &    \Third{0.9051}  &   \First{0.8713}   &   \First{0.9443}   &  &  \First{0.0159}    &   \First{0.9248}   &  \First{0.8828}    &   \First{0.9529}  &  &   \Third{50.87}    & \Third{112.63}     \\ \hline
\end{tabular}
}
\end{table}

\begin{figure}[!h]
\centering
\includegraphics[width=1.0\linewidth]{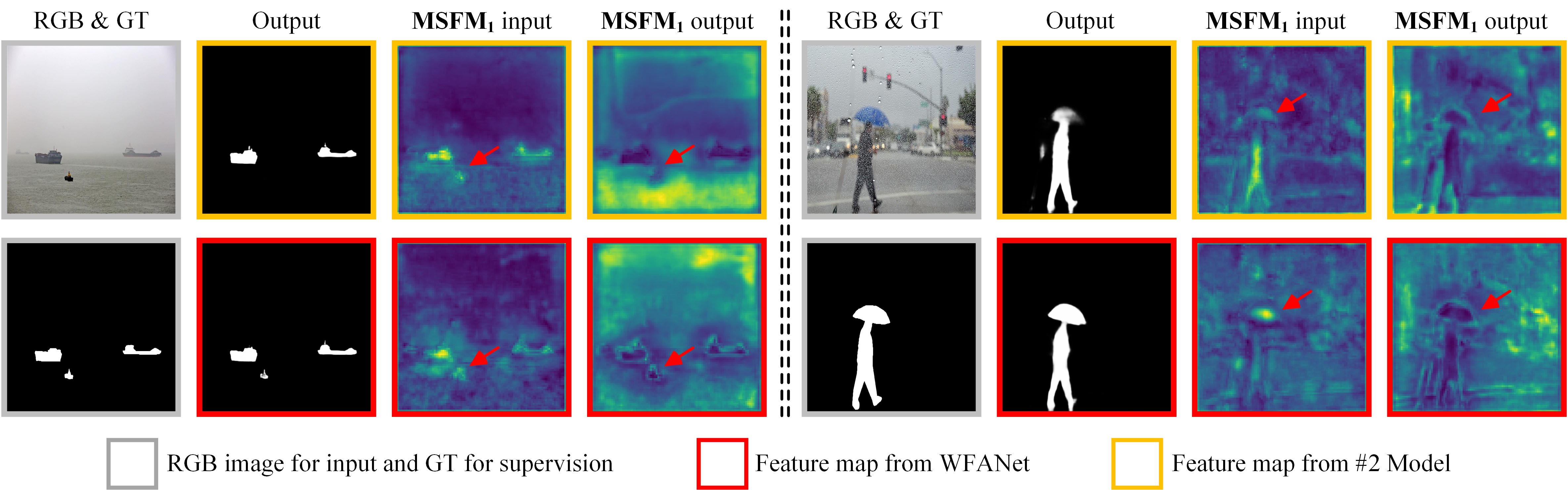}
\caption{Comparison of Input and Output Features of MSFM Between WFNAet and \#2 Model.}
\label{mid_feature}
\end{figure}  

To intuitively demonstrate the advantages of the dual-branch model, we present feature maps of semantic features before and after optimization via the weather prediction branch output, as shown in Fig.~\ref{mid_feature}. Comparing input and output feature maps of MSFM at the same scale between WFANet and \#2 Model, WFANet's output exhibits clearer edge details, better aligning with salient objects. Thus, WFANet enables more complete segmentation of salient objects under noise interference.

\begin{figure}[!h]
\centering
\includegraphics[width=0.9\linewidth]{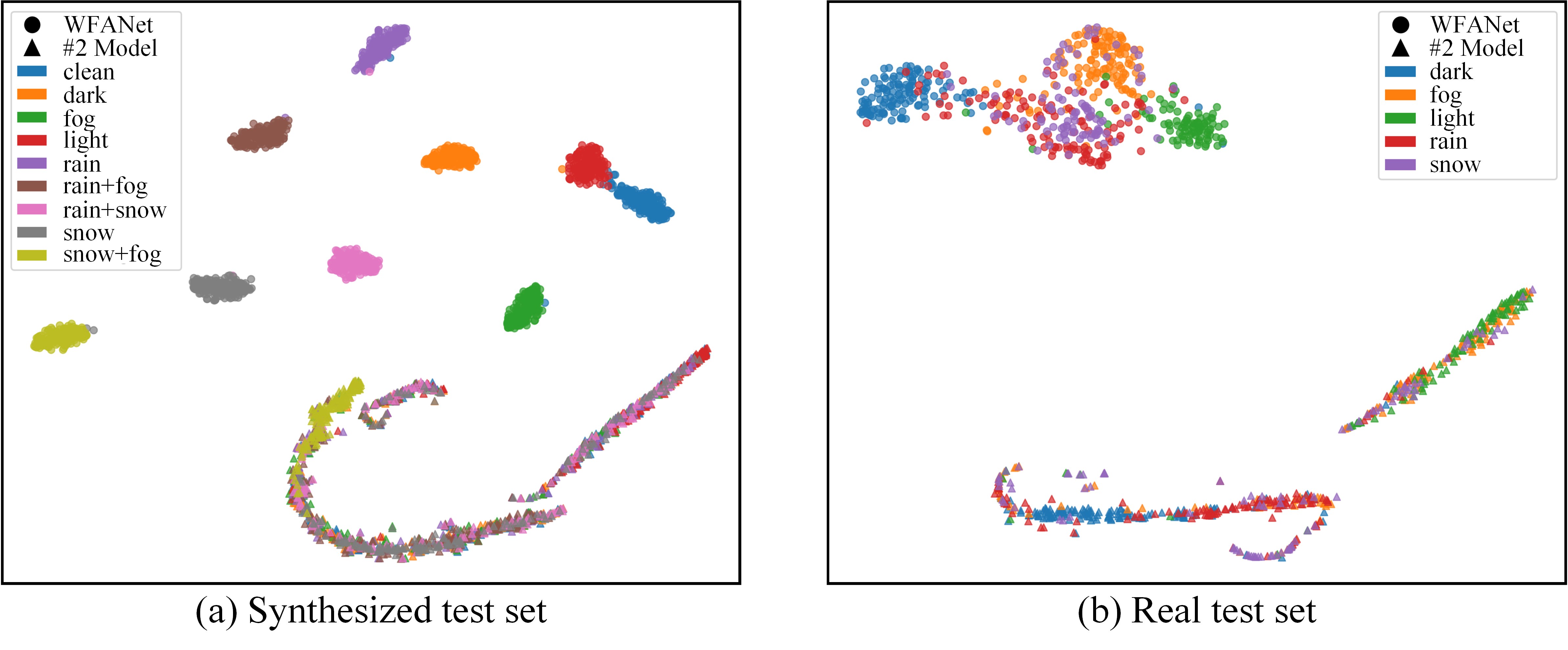}
\caption{Visualization of WPM output features for \#2 Model and WFANet using t-SNE.}
\label{t-sne1}
\end{figure}  

To verify that the supervised weather prediction branch can learn noise-related features, we visualize the output features of the WPM from \#2 Model and WFANet using t-SNE, as shown in Fig.~\ref{t-sne1}. Subfigure~(a) presents comparison results on the synthesized test set, and (b) on the real test set. Compared with \#2 Model, the WPM-derived features of WFANet exhibit a more discrete distribution, where samples with same noise types show stronger aggregation. This confirms that WFANet’s supervised optimization of the weather prediction branch enables the WPM to learn noise-related features, thereby enhancing the model's performance. Notably, the model’s ability to distinguish rain and snow noises requires further improvement, primarily due to their high similarity in real scenarios.

\begin{figure}[!h]
\centering
\includegraphics[width=0.7\linewidth]{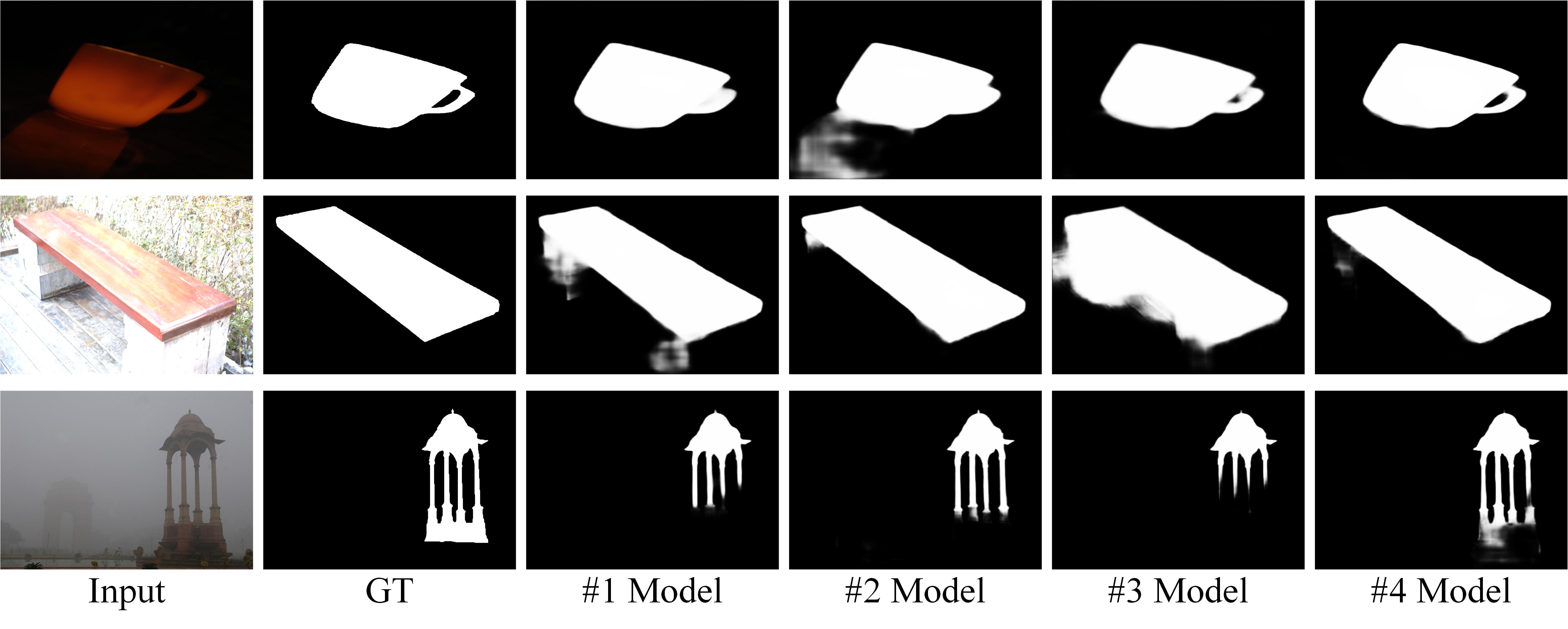}
\caption{Visual results obtained by the model using different structures.}
\label{Quantitative evaluation structures}
\end{figure}

Visualization results of the variant models are illustrated in Fig.~\ref{Quantitative evaluation structures}. It can be observed that, in contrast to the variant models, WFANet is capable of mitigating the interference of weather-related noise and accurately segmenting salient objects from input images. Collectively, both quantitative and qualitative results fully validate the superiority of the dual-branch architecture in WFANet.

\textbf{2) Backbone Network Compatibility.}
To explore the impact of different backbone networks on the performance of WFANet, we conducted ablation studies by replacing the backbone networks in both Branch-1 (saliency detection branch) and Branch-2 (weather prediction branch). The results are summarized in Table~\ref{backbone result}.

\begin{table}[!h]
\centering
\caption{Ablation studies of our model under different backbone networks.}
\label{backbone result}
\resizebox{1.0\linewidth}{!}{
\begin{tabular}{lcccccccccccccc}
\hline
\multirow{2}{*}{Branch-1} & \multirow{2}{*}{Branch-1} &  & \multicolumn{4}{c}{Synthesized test set} &  & \multicolumn{4}{c}{Real test set} &  & \multicolumn{2}{c}{Computational costs} \\ \cline{4-7} \cline{9-12} \cline{14-15} 
&         &  & $MAE\downarrow$ & $S\uparrow$ & $F_{\beta}^{mean}\uparrow$ & $E_{}^{mean}\uparrow$   &  & $MAE\downarrow$ & $S\uparrow$ & $F_{\beta}^{mean}\uparrow$ & $E_{}^{mean}\uparrow$  &  & $Parmas$(M)      & $MACs$(G)       \\ \hline
\multirow{5}{*}{PVTv2-b}  & ResNet18                  &  &  \Second{0.0229}    &    0.9051  &    \Third{0.8713} &   \Second{0.9443}  &  &   \Second{0.0159}   &   \First{0.9248}   &  \First{0.8828}    &  \Second{0.9529}   &  &   \Second{50.87}             &     \Second{112.63}   \\
& ResNet50                  &  &   0.0233   &  \Third{0.9059}    &  \First{0.8716}   &  \Third{0.9439}   &  &  0.0193    &   0.9114   &    0.8616  &  0.9390   &  &     66.74    &  \Third{119.92} \\
& VGG16                     &  & \Third{0.0230}     &  \Second{0.9076}    &  0.8707   &   0.9414  &  &  \First{0.0156}    &  \Second{0.9248}    &   \Second{0.8790}   &   \First{0.9542}  &  &    \Third{54.41}   &   152.38  \\
& Swin-B                    &  &   0.0990   &  0.5961    &   0.4303  &   0.5304  &  &  0.0902    &   0.5885   &   0.3797   &   0.5156  &  &    127.56  &   152.10   \\ 
& MobileNet                    &  &   \First{0.0220}   &  \First{0.9087}    &   \Second{0.8715}  &   \First{0.9453}  &  &  \Third{0.0167}    &   \Third{0.9157}   &  \Third{0.8696}   &   \Third{0.9464} &  &   \First{39.61}   &   \First{107.30}   \\ 
\hline\hline
ResNet50                  & \multirow{4}{*}{ResNet18} &  & \Third{0.0361}     &  \Third{0.8661}    &  \Third{0.8137}   &   \Third{0.9019}  &  &    \Third{0.0246} &   \Third{0.8936}   &  \Third{0.8390}    &  \Third{0.9277}   &  &    \Second{50.25}   &    \Second{114.15}     \\
VGG16                     &                           &  &  0.0379    &   0.8492   &   0.8018  &  0.8806   &  &    0.0276  &   0.8730   &   0.8226   &  0.8986   &  &   \First{41.28}     &   547.57   \\
Swin-B                    &                           &  &  \First{0.0216}    &   \Second{0.9044}   & \First{0.8744}    & \First{0.9471}    &  &   \First{0.0155}   &    \Second{0.9148}  &   \Second{0.8757}   & \Second{0.9466}    &  &   112.93    &  \Third{135.39}   \\
PVTv2-b                   &                           &  &    \Second{0.0229}  &    \First{0.9051}  & \Second{0.8713}    &  \Second{0.9443}   &  &    \Second{0.0159}  &   \First{0.9248}   &  \First{0.8828}    &  \First{0.9529}   &  &  \Third{50.87}   & \First{112.63}    \\ \hline
\end{tabular}
}
\end{table}

\begin{figure}[!h]
\centering
\includegraphics[width=0.7\linewidth]{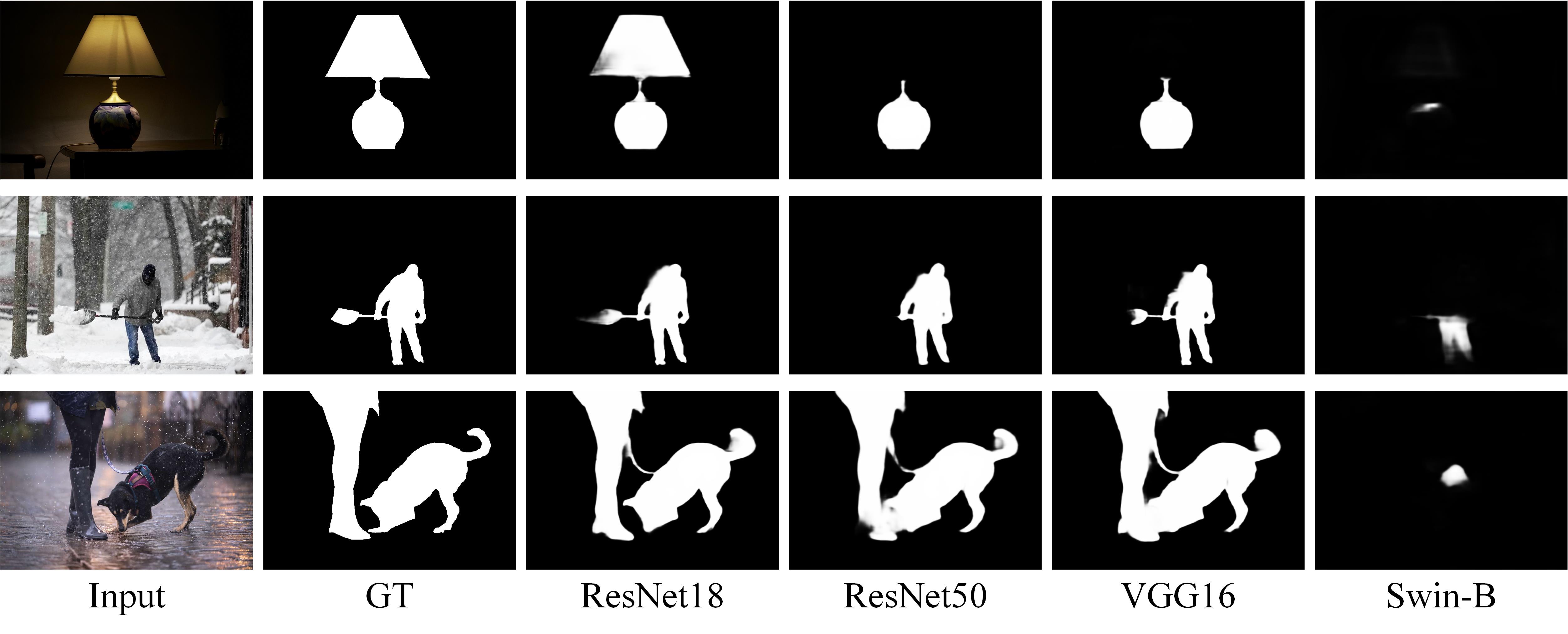}
\caption{Visual results obtained by the weather prediction branch using different backbones.}
\label{Quantitative evaluation WPBbackbone}
\end{figure}

\begin{figure}[!h]
\centering
\includegraphics[width=0.7\linewidth]{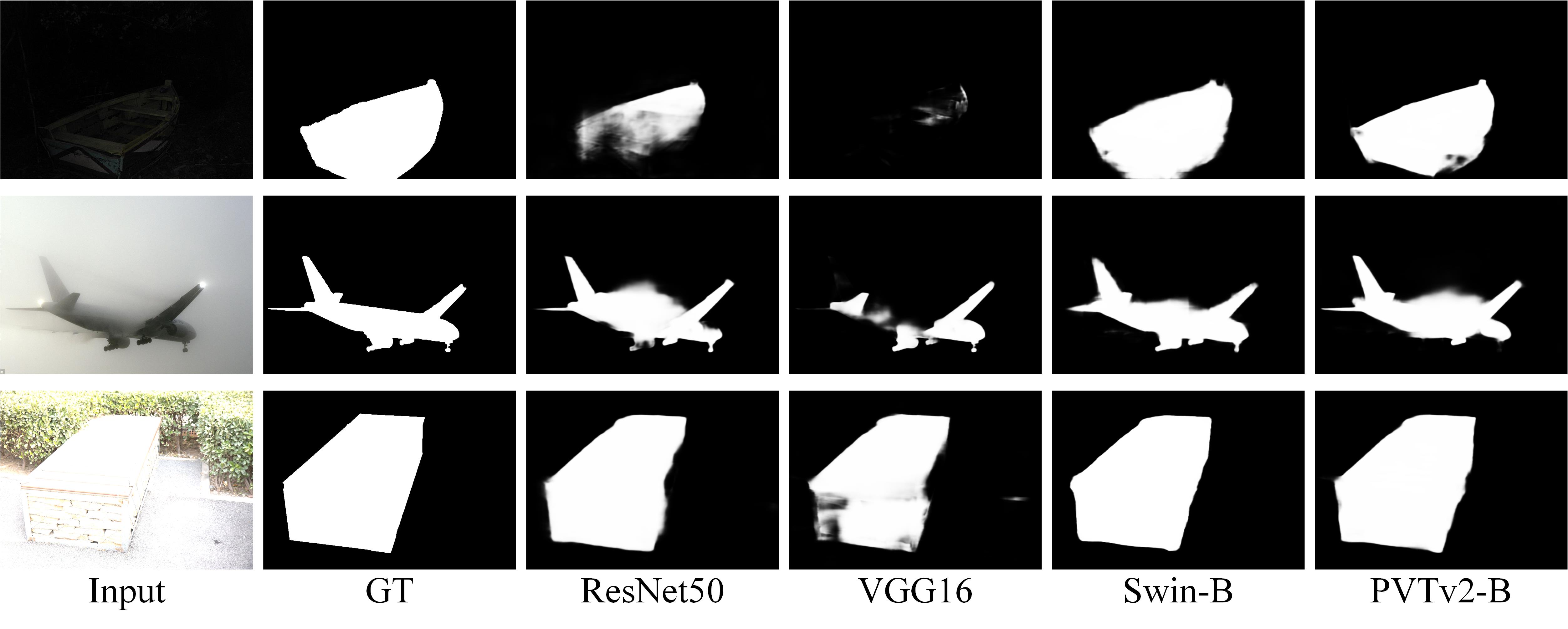}
\caption{Visual results obtained by the saliency detection branch using different backbones.}
\label{Quantitative evaluation SDBbackbone}
\end{figure}

The upper part maintains the Branch-1 fixed with a PVTv2-b backbone while varying the backbone of the Branch-2 among ResNet18, ResNet50, VGG16, Swin-B and MobileNet. Light-weight ResNet18 and MobileNet yield performance generally comparable to the heavier ResNet50, particularly in $S$ and $F_{\beta}$ metrics, suggesting that a simpler network is sufficient and potentially optimal. 
Note that the combination of PVTv2-b and Swin-B exhibits a notable decline in effectiveness. We attribute it to the large parameter sizes of both, which makes it difficult to achieve effective feature fusion using a unified training strategy. 
The lower part fixes Branch-2 to ResNet18 while varying the Branch-1 backbone (ResNet50, VGG16, Swin-B and PVTv2-b).
This test distinctly highlights the crucial role of the saliency branch backbone.
The PVTv2-b and Swin-B backbones deliver the top two quantitative performance across all metrics. Its superior feature extraction capability boosts the model's ability to detect salient objects accurately under varying weather conditions. The combination of PVTv2-b for Branch-1 and ResNet18 for Branch-2 represents an effective balance, yielding strong performance across the evaluated quantitative results. 
Fig.~\ref{Quantitative evaluation WPBbackbone} and Fig.~\ref{Quantitative evaluation SDBbackbone} present the corresponding visualization results, which further validate the performance superiority of the PVTv2-b and ResNet18 combination.

\textbf{3) Basic Unit Settings for WPM and MSFM.}
We also investigate the impact of the number of basic units ($CNR$) within the weather prediction module and multi-source fusion module. Experimental results (Tables~\ref{WPM ablation result} and~\ref{MSFM ablation result}) demonstrate that incrementally increasing the number of units in both modules consistently enhances detection accuracy, with optimal performance achieved when three units are deployed in each module. As the number of basic units is further increased, the model performance tends to stabilize while the computational cost continues to rise. Consequently, WFANet adopts three $CNR$ units in both WPM and MSFM as the default configuration.

\begin{table}[!h]
\centering
\caption{Ablation studies of WPM with varying numbers of $CNR$.}
\label{WPM ablation result}
\resizebox{1.0\linewidth}{!}{
\begin{tabular}{lccccccccccccc}
\hline
\multirow{2}{*}{Number} &  & \multicolumn{4}{c}{Synthesized test set} &  & \multicolumn{4}{c}{Real test set} &  & \multicolumn{2}{c}{Computational costs} \\ \cline{3-6} \cline{8-11} \cline{13-14} 
&  & $MAE\downarrow$   & $S\uparrow$       & $F_{\beta}^{mean}\uparrow$       & $E_{}^{mean}\uparrow$   &  & $MAE\downarrow$   & $S\uparrow$       & $F_{\beta}^{mean}\uparrow$       & $E_{}^{mean}\uparrow$   &  & $Params$(M)           & $MACs$(G)               \\ \hline
1                       &  &   0.0230   &   \First{0.9073}   &   0.8684  &  0.9401   &  &   0.0175       &  0.9167       &    0.8654     &   0.9449      &  &     \First{49.69}                &    \First{112.46}         \\
2                       &  &   \First{0.0223}   &  \Second{0.9070}    & \Second{0.8705}    &   \Second{0.9430}  &  &   0.0162        &    \Second{0.9231}     &   \Second{0.8811}      &    \Third{0.9506}     &  &        \Second{50.28}             &  \Second{112.55}    \\
3                       &  &   \Second{0.0229}   &   0.9051   &   \First{0.8713}  &  \First{0.9443}   &  &    \First{0.0159}       &    \First{0.9248}     & \First{0.8828}        &    \Second{0.9529}     &  &     \Third{50.87}                &    \Third{112.63}           \\
4                       &  &   0.0234   &  0.9059    &   0.8674  & \Third{0.9429}    &  &    \Second{0.0160}      &  0.9225       &  0.8739       &     \First{0.9543}    &  &       51.46              &    112.72           \\
5                       &  &   \Third{0.0229}   &  \Third{0.9067}    &  \Third{0.8691}   &   0.9427  &  &     \Third{0.0161}        &  \Third{0.9228}       &  \Third{0.8775}       &     0.9505    &  &    52.05                 &     112.80           \\ \hline
\end{tabular}
}
\end{table}

\begin{table}[!h]
\centering
\caption{Ablation studies of MSFM with varying numbers of $CNR$.}
\label{MSFM ablation result}
\resizebox{1.0\linewidth}{!}{
\begin{tabular}{lccccccccccccc}
\hline
\multirow{2}{*}{Number} &  & \multicolumn{4}{c}{Synthesized test set} &  & \multicolumn{4}{c}{Real test set} &  & \multicolumn{2}{c}{Computational costs} \\ \cline{3-6} \cline{8-11} \cline{13-14} 
&  & $MAE\downarrow$   & $S\uparrow$       & $F_{\beta}^{mean}\uparrow$       & $E_{}^{mean}\uparrow$   &  & $MAE\downarrow$   & $S\uparrow$       & $F_{\beta}^{mean}\uparrow$       & $E_{}^{mean}\uparrow$   &  & $Params$(M)           & $MACs$(G)               \\ \hline
1                       &  &   \Third{0.0235}   &   \Third{0.9048}   &   0.8693  &  \Third{0.9424}   &  &   0.0195        &    0.9135     &   0.8611      &    0.9380     &  &        \First{47.33}       &    \First{69.26}              \\
2                       &  &   \First{0.0218}   &  \First{0.9090}    & \Second{0.8699}    &   \First{0.9459}  &  &    0.0178        &    0.9161     &    0.8701     &   \Third{0.9449}      &  &      \Second{49.10}        &   \Second{90.95}          \\
3                       &  &   \Second{0.0229}   &   \Second{0.9051}   &   \First{0.8713}  &  \Second{0.9443}   &  &    \First{0.0159}        &   \First{0.9248}      & \First{0.8828}        &    \First{0.9529}     &  &     \Third{50.87}          &   \Third{112.63}               \\
4                       &  &   0.0240   &  0.9019    &   0.8628  & 0.9402    &  &    \Third{0.0172}          &   \Third{0.9198}      &  \Third{0.8724}       &   0.9435      &  &          52.64           &    134.31               \\
5                       &  &   0.0240   &  0.9045    &  \Third{0.8698}   &   0.9421  &  &     \Second{0.0170}       &   \Second{0.9223}      &    \Second{0.8770}     &    \Second{0.9486}     &  &        54.41             &   156.00        \\ \hline
\end{tabular}
}
\end{table}

\subsection{Failure Cases and Analysis}\label{Failure Cases and Analysis}
\begin{figure}[!h]
\centering
\includegraphics[width=0.7\linewidth]{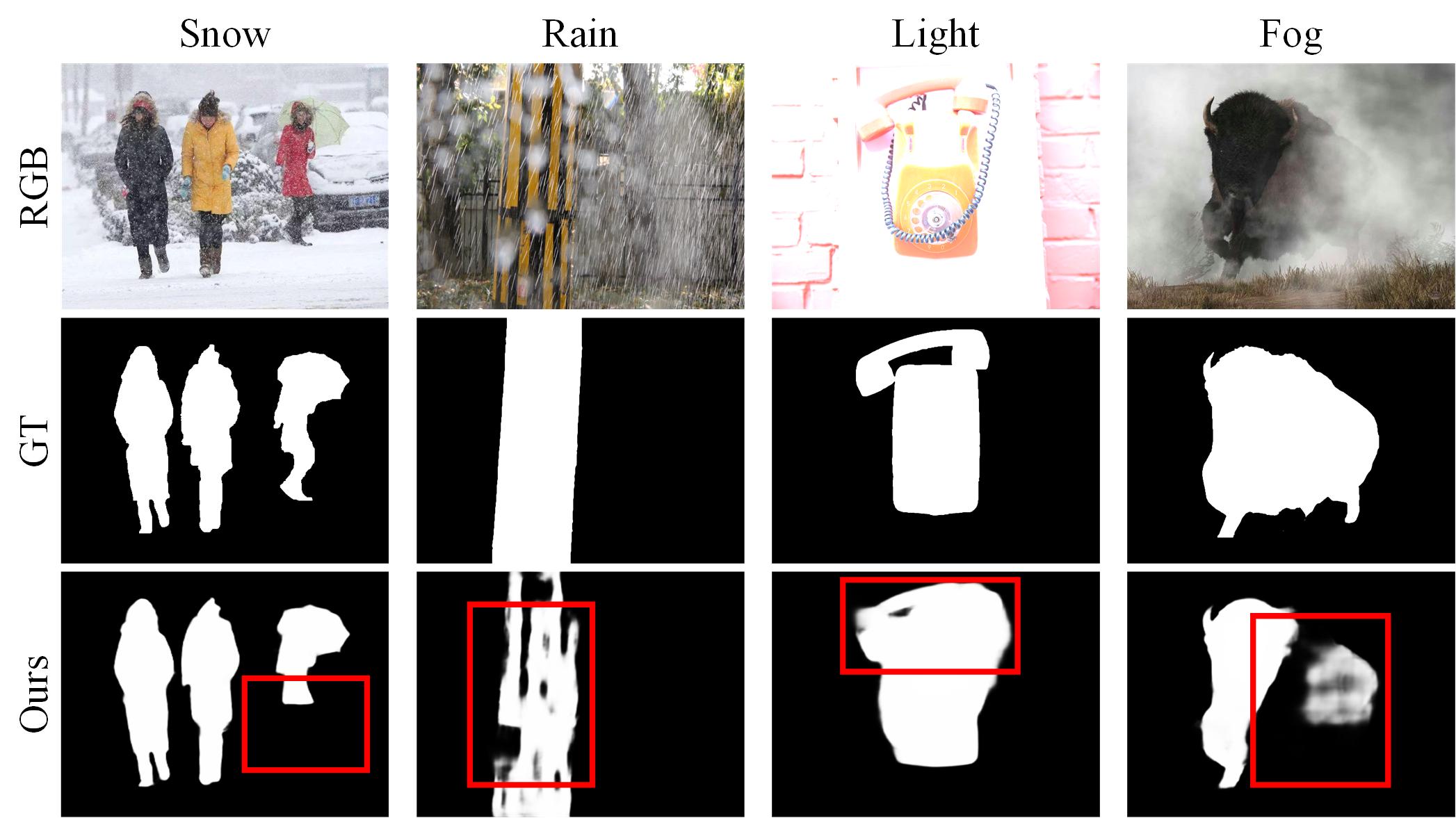}
\caption{Some failure examples.}
\label{failure cases}
\end{figure}

Extensive experiments have proven that WFANet is robust against various weather noise and obtains precise predictions results. However, in certain extreme scenarios, WFANet still struggles to achieve perfect results. As shown in Fig.~\ref{failure cases}, we present four challenging examples. Rain and snow scenarios contain dense visual noise, leading to misclassifications of salient regions and errors in boundary detection. Light and fog scenarios visually compromise target integrity, making it challenging for the model to detect boundaries in extreme areas. In future work, we will focus on a detailed analysis of the differences between various weather noises in both the spatial and frequency domains. Additionally, multi-modal data will be leveraged to mitigate noise interference, thereby enhancing accuracy and robustness in complex real-world scenarios.

\section{Conclusion}\label{sec:Conclusion}
This paper presents WXSOD, a large-scale dataset for salient object detection in adverse weather conditions. WXSOD contains 14,945 RGB images with diverse weather noise, along with pixel-wise ground truth annotations and weather labels. Its dual test set (synthetic noise and real-world noise) ensure the establishment of a rigorous benchmark for generalization assessment. Building upon WXSOD, we propose an efficient baseline model named WFANet, which employs a fully supervised two-branch architecture. The first branch focuses on predicting weather conditions by extracting noise-related features, while the second branch integrates these noise-related features with semantic features for saliency detection. Extensive experiments against 17 state-of-the-art SOD methods show WFANet’s superior performance on WXSOD, highlighting its effectiveness in adverse weather conditions.

As an emerging sub-field, salient object detection in adverse weather conditions still has significant room for improvement. 
While WFANet serves as a reliable baseline, its computational complexity may limit deployment in resource-constrained scenarios (\textit{e.g.}, edge devices). Future work will focus on developing lightweight variants of WFANet that maintain high accuracy while reducing computational costs.
Additionally, research into adaptive mechanisms that dynamically modulate feature aggregation based on weather severity remains an open challenge. 


\bibliographystyle{elsarticle-num} 
\bibliography{ref}

\begin{thebibliography}{10}
\expandafter\ifx\csname url\endcsname\relax
  \def\url#1{\texttt{#1}}\fi
\expandafter\ifx\csname urlprefix\endcsname\relax\def\urlprefix{URL }\fi
\expandafter\ifx\csname href\endcsname\relax
  \def\href#1#2{#2} \def\path#1{#1}\fi

\bibitem{cong2018review}
R.~Cong, J.~Lei, H.~Fu, M.-M. Cheng, W.~Lin, Q.~Huang, Review of visual saliency detection with comprehensive information, IEEE Transactions on circuits and Systems for Video Technology 29~(10) (2018) 2941--2959.

\bibitem{wei2020label}
J.~Wei, S.~Wang, Z.~Wu, C.~Su, Q.~Huang, Q.~Tian, Label decoupling framework for salient object detection, in: Proceedings of the IEEE/CVF conference on computer vision and pattern recognition, 2020, pp. 13025--13034.

\bibitem{fang2021ibnet}
X.~Fang, J.~Zhu, R.~Zhang, X.~Shao, H.~Wang, Ibnet: Interactive branch network for salient object detection, Neurocomputing 465 (2021) 574--583.

\bibitem{feng2020residual}
M.~Feng, H.~Lu, Y.~Yu, Residual learning for salient object detection, IEEE Transactions on Image Processing 29 (2020) 4696--4708.

\bibitem{wu2019cascaded}
Z.~Wu, L.~Su, Q.~Huang, Cascaded partial decoder for fast and accurate salient object detection, in: Proceedings of the IEEE/CVF conference on computer vision and pattern recognition, 2019, pp. 3907--3916.

\bibitem{li2018contour}
X.~Li, F.~Yang, H.~Cheng, W.~Liu, D.~Shen, Contour knowledge transfer for salient object detection, in: Proceedings of the european conference on computer vision (ECCV), 2018, pp. 355--370.

\bibitem{liang2024meanet}
B.~Liang, H.~Luo, Meanet: An effective and lightweight solution for salient object detection in optical remote sensing images, Expert Systems with Applications 238 (2024) 121778.

\bibitem{zamir2022restormer}
S.~W. Zamir, A.~Arora, S.~Khan, M.~Hayat, F.~S. Khan, M.-H. Yang, Restormer: Efficient transformer for high-resolution image restoration, in: Proceedings of the IEEE/CVF conference on computer vision and pattern recognition, 2022, pp. 5728--5739.

\bibitem{jiang2023dawn}
K.~Jiang, W.~Liu, Z.~Wang, X.~Zhong, J.~Jiang, C.-W. Lin, Dawn: Direction-aware attention wavelet network for image deraining, in: Proceedings of the 31st ACM international conference on multimedia, 2023, pp. 7065--7074.

\bibitem{sun2023event}
S.~Sun, W.~Ren, J.~Li, K.~Zhang, M.~Liang, X.~Cao, Event-aware video deraining via multi-patch progressive learning, IEEE Transactions on Image Processing 32 (2023) 3040--3053.

\bibitem{li2022all}
B.~Li, X.~Liu, P.~Hu, Z.~Wu, J.~Lv, X.~Peng, All-in-one image restoration for unknown corruption, in: Proceedings of the IEEE/CVF conference on computer vision and pattern recognition, 2022, pp. 17452--17462.

\bibitem{potlapalli2024promptir}
V.~Potlapalli, S.~W. Zamir, S.~H. Khan, F.~Shahbaz~Khan, Promptir: Prompting for all-in-one image restoration, Advances in Neural Information Processing Systems 36 (2024).

\bibitem{ai2024multimodal}
Y.~Ai, H.~Huang, X.~Zhou, J.~Wang, R.~He, Multimodal prompt perceiver: Empower adaptiveness generalizability and fidelity for all-in-one image restoration, in: Proceedings of the IEEE/CVF Conference on Computer Vision and Pattern Recognition, 2024, pp. 25432--25444.

\bibitem{yang2013saliency}
C.~Yang, L.~Zhang, H.~Lu, X.~Ruan, M.-H. Yang, Saliency detection via graph-based manifold ranking, in: Proceedings of the IEEE conference on computer vision and pattern recognition, 2013, pp. 3166--3173.

\bibitem{wang2017learning}
L.~Wang, H.~Lu, Y.~Wang, M.~Feng, D.~Wang, B.~Yin, X.~Ruan, Learning to detect salient objects with image-level supervision, in: Proceedings of the IEEE conference on computer vision and pattern recognition, 2017, pp. 136--145.

\bibitem{li2014secrets}
Y.~Li, X.~Hou, C.~Koch, J.~M. Rehg, A.~L. Yuille, The secrets of salient object segmentation, in: Proceedings of the IEEE conference on computer vision and pattern recognition, 2014, pp. 280--287.

\bibitem{li2019nested}
C.~Li, R.~Cong, J.~Hou, S.~Zhang, Y.~Qian, S.~Kwong, Nested network with two-stream pyramid for salient object detection in optical remote sensing images, IEEE Transactions on Geoscience and Remote Sensing 57~(11) (2019) 9156--9166.

\bibitem{zhang2020dense}
Q.~Zhang, R.~Cong, C.~Li, M.-M. Cheng, Y.~Fang, X.~Cao, Y.~Zhao, S.~Kwong, Dense attention fluid network for salient object detection in optical remote sensing images, IEEE Transactions on Image Processing 30 (2020) 1305--1317.

\bibitem{tu2022rgbt}
Z.~Tu, Y.~Ma, Z.~Li, C.~Li, J.~Xu, Y.~Liu, Rgbt salient object detection: A large-scale dataset and benchmark, IEEE Transactions on Multimedia 25 (2022) 4163--4176.

\bibitem{wang2024alignment}
K.~Wang, D.~Lin, C.~Li, Z.~Tu, B.~Luo, Alignment-free rgbt salient object detection: Semantics-guided asymmetric correlation network and a unified benchmark, IEEE Transactions on Multimedia 26 (2024) 10692--10707.

\bibitem{song2023modality}
K.~Song, H.~Wen, X.~Xue, L.~Huang, Y.~Ji, Y.~Yan, Modality registration and object search framework for uav-based unregistered rgb-t image salient object detection, IEEE Transactions on Geoscience and Remote Sensing 61 (2023) 1--15.

\bibitem{jung2020imgaug}
A.~B. Jung, K.~Wada, J.~Crall, S.~Tanaka, J.~Graving, C.~Reinders, S.~Yadav, J.~Banerjee, G.~Vecsei, A.~Kraft, et~al., imgaug, GitHub: San Francisco, CA, USA (2020).

\bibitem{jian2017ouc}
M.~Jian, Q.~Qi, J.~Dong, Y.~Yin, W.~Zhang, K.-M. Lam, The ouc-vision large-scale underwater image database, in: 2017 IEEE International Conference on Multimedia and Expo (ICME), IEEE, 2017, pp. 1297--1302.

\bibitem{islam2022svam}
M.~J. Islam, R.~Wang, J.~Sattar, Svam: Saliency-guided visual attention modeling by autonomous underwater robot, in: Proceedings of Robotics: Science and Systems, 2022.

\bibitem{hong2023usod10k}
L.~Hong, X.~Wang, G.~Zhang, M.~Zhao, Usod10k: a new benchmark dataset for underwater salient object detection, IEEE transactions on image processing (2023).

\bibitem{peng2014rgbd}
H.~Peng, B.~Li, W.~Xiong, W.~Hu, R.~Ji, Rgbd salient object detection: A benchmark and algorithms, in: Computer Vision--ECCV 2014: 13th European Conference, Zurich, Switzerland, September 6-12, 2014, Proceedings, Part III 13, Springer, 2014, pp. 92--109.

\bibitem{cheng2014depth}
Y.~Cheng, H.~Fu, X.~Wei, J.~Xiao, X.~Cao, Depth enhanced saliency detection method, in: Proceedings of international conference on internet multimedia computing and service, 2014, pp. 23--27.

\bibitem{zhu2017three}
C.~Zhu, G.~Li, A three-pathway psychobiological framework of salient object detection using stereoscopic technology, in: Proceedings of the IEEE international conference on computer vision workshops, 2017, pp. 3008--3014.

\bibitem{niu2012leveraging}
Y.~Niu, Y.~Geng, X.~Li, F.~Liu, Leveraging stereopsis for saliency analysis, in: 2012 IEEE conference on computer vision and pattern recognition, IEEE, 2012, pp. 454--461.

\bibitem{wang2018rgb}
G.~Wang, C.~Li, Y.~Ma, A.~Zheng, J.~Tang, B.~Luo, Rgb-t saliency detection benchmark: Dataset, baselines, analysis and a novel approach, in: Image and Graphics Technologies and Applications: 13th Conference on Image and Graphics Technologies and Applications, IGTA 2018, Beijing, China, April 8--10, 2018, Revised Selected Papers 13, Springer, 2018, pp. 359--369.

\bibitem{tu2019rgb}
Z.~Tu, T.~Xia, C.~Li, X.~Wang, Y.~Ma, J.~Tang, Rgb-t image saliency detection via collaborative graph learning, IEEE Transactions on Multimedia 22~(1) (2019) 160--173.

\bibitem{song2022novel}
K.~Song, J.~Wang, Y.~Bao, L.~Huang, Y.~Yan, A novel visible-depth-thermal image dataset of salient object detection for robotic visual perception, IEEE/ASME Transactions on Mechatronics 28~(3) (2022) 1558--1569.

\bibitem{yu2024degradation}
N.~Yu, J.~Wang, H.~Shi, Z.~Zhang, Y.~Han, Degradation-removed multiscale fusion for low-light salient object detection, Pattern Recognition (2024) 110650.

\bibitem{achanta2009frequency}
R.~Achanta, S.~Hemami, F.~Estrada, S.~Susstrunk, Frequency-tuned salient region detection, in: 2009 IEEE conference on computer vision and pattern recognition, IEEE, 2009, pp. 1597--1604.

\bibitem{peng2016salient}
H.~Peng, B.~Li, H.~Ling, W.~Hu, W.~Xiong, S.~J. Maybank, Salient object detection via structured matrix decomposition, IEEE transactions on pattern analysis and machine intelligence 39~(4) (2016) 818--832.

\bibitem{zhang2015minimum}
J.~Zhang, S.~Sclaroff, Z.~Lin, X.~Shen, B.~Price, R.~Mech, Minimum barrier salient object detection at 80 fps, in: Proceedings of the IEEE international conference on computer vision, 2015, pp. 1404--1412.

\bibitem{chen2024sdpl}
Q.~Chen, T.~Wang, Z.~Yang, H.~Li, R.~Lu, Y.~Sun, B.~Zheng, C.~Yan, Sdpl: Shifting-dense partition learning for uav-view geo-localization, IEEE Transactions on Circuits and Systems for Video Technology 34~(11) (2024) 11810--11824.

\bibitem{zheng2021learning}
B.~Zheng, S.~Yuan, C.~Yan, X.~Tian, J.~Zhang, Y.~Sun, L.~Liu, A.~Leonardis, G.~Slabaugh, Learning frequency domain priors for image demoireing, IEEE Transactions on Pattern Analysis and Machine Intelligence 44~(11) (2021) 7705--7717.

\bibitem{he2025samba}
J.~He, K.~Fu, X.~Liu, Q.~Zhao, Samba: A unified mamba-based framework for general salient object detection, in: Proceedings of the Computer Vision and Pattern Recognition Conference, 2025, pp. 25314--25324.

\bibitem{he2016deep}
K.~He, X.~Zhang, S.~Ren, J.~Sun, Deep residual learning for image recognition, in: Proceedings of the IEEE conference on computer vision and pattern recognition, 2016, pp. 770--778.

\bibitem{chen2018reverse}
S.~Chen, X.~Tan, B.~Wang, X.~Hu, Reverse attention for salient object detection, in: Proceedings of the European conference on computer vision (ECCV), 2018, pp. 234--250.

\bibitem{song2023salient}
X.~Song, F.~Guo, L.~Zhang, X.~Lu, X.~Hei, Salient object detection with dual-branch stepwise feature fusion and edge refinement, IEEE Transactions on Circuits and Systems for Video Technology (2023).

\bibitem{wu2024misclassification}
Z.~Wu, Y.~Xu, J.~Yang, X.~Li, Misclassification in weakly supervised object detection, IEEE Transactions on Image Processing 33 (2024) 3413--3427.

\bibitem{wu2025weakly}
Z.~Wu, Y.~Xu, J.~Yang, D.~Zhang, Weakly supervised salient object detection with oversize bounding box: Z. wu et al., International Journal of Computer Vision (2025) 1--20.

\bibitem{li2025ifa}
M.~Li, H.~Zhang, K.~Cai, W.~Pedrycz, D.~Miao, Y.~Gao, Ifa: Illumination-aware feature aggregation model for salient object detection, Pattern Recognition (2025) 112118.

\bibitem{zhu2025dc}
J.~Zhu, X.~Qin, A.~Elsaddik, Dc-net: Divide-and-conquer for salient object detection, Pattern Recognition 157 (2025) 110903.

\bibitem{liu2020dynamic}
J.-J. Liu, Q.~Hou, M.-M. Cheng, Dynamic feature integration for simultaneous detection of salient object, edge, and skeleton, IEEE Transactions on Image Processing 29 (2020) 8652--8667.

\bibitem{zhou2020interactive}
H.~Zhou, X.~Xie, J.-H. Lai, Z.~Chen, L.~Yang, Interactive two-stream decoder for accurate and fast saliency detection, in: Proceedings of the IEEE/CVF conference on computer vision and pattern recognition, 2020, pp. 9141--9150.

\bibitem{wen2021dynamic}
H.~Wen, C.~Yan, X.~Zhou, R.~Cong, Y.~Sun, B.~Zheng, J.~Zhang, Y.~Bao, G.~Ding, Dynamic selective network for rgb-d salient object detection, IEEE Transactions on Image Processing 30 (2021) 9179--9192.

\bibitem{zhu2024cmignet}
H.~Zhu, J.~Ni, X.~Yang, L.~Zhang, Cmignet: Cross-modal inverse guidance network for rgb-depth salient object detection, Pattern Recognition 155 (2024) 110693.

\bibitem{yang2025mitigating}
Y.~Yang, N.~Huang, Q.~Zhang, J.~Han, Mitigating fusion bias for rgb-d salient object detection, Pattern Recognition (2025) 112135.

\bibitem{zhong2025lesod}
M.~Zhong, J.~Sun, F.~Wang, F.~Sun, Lesod: Lightweight and efficient network for rgb-d salient object detection, Pattern Recognition (2025) 112103.

\bibitem{gao2023depth}
L.~Gao, B.~Liu, P.~Fu, M.~Xu, Depth-aware inverted refinement network for rgb-d salient object detection, Neurocomputing 518 (2023) 507--522.

\bibitem{cong2022does}
R.~Cong, K.~Zhang, C.~Zhang, F.~Zheng, Y.~Zhao, Q.~Huang, S.~Kwong, Does thermal really always matter for rgb-t salient object detection?, IEEE Transactions on Multimedia 25 (2022) 6971--6982.

\bibitem{zhou2024frequency}
H.~Zhou, C.~Tian, Z.~Zhang, C.~Li, Y.~Xie, Z.~Li, Frequency-aware feature aggregation network with dual-task consistency for rgb-t salient object detection, Pattern Recognition 146 (2024) 110043.

\bibitem{wang2024learning}
K.~Wang, Z.~Tu, C.~Li, C.~Zhang, B.~Luo, Learning adaptive fusion bank for multi-modal salient object detection, IEEE Transactions on Circuits and Systems for Video Technology 34~(8) (2024) 7344--7358.

\bibitem{zhou2025deformation}
H.~Zhou, Z.~Zhang, C.~Li, C.~Tian, Y.~Xie, Z.~Li, X.-J. Wu, Deformation-resilient multigranularity learning for unaligned rgb--t semantic segmentation, IEEE Transactions on Neural Networks and Learning Systems (2025).

\bibitem{luo2024dynamic}
Y.~Luo, F.~Shao, B.~Mu, H.~Chen, Z.~Li, Q.~Jiang, Dynamic weighted fusion and progressive refinement network for visible-depth-thermal salient object detection, IEEE Transactions on Circuits and Systems for Video Technology (2024).

\bibitem{wan2024tmnet}
B.~Wan, X.~Zhou, Y.~Sun, Z.~Zhu, H.~Wang, C.~Yan, et~al., Tmnet: Triple-modal interaction encoder and multi-scale fusion decoder network for vdt salient object detection, Pattern Recognition 147 (2024) 110074.

\bibitem{bao2024quality}
L.~Bao, X.~Zhou, X.~Lu, Y.~Sun, H.~Yin, Z.~Hu, J.~Zhang, C.~Yan, Quality-aware selective fusion network for vdt salient object detection, IEEE Transactions on Image Processing (2024).

\bibitem{wang2025unified}
K.~Wang, Z.~Tu, C.~Li, Z.~Liu, B.~Luo, Unified-modal salient object detection via adaptive prompt learning, IEEE Transactions on Circuits and Systems for Video Technology (2025).

\bibitem{shi2015hierarchical}
J.~Shi, Q.~Yan, L.~Xu, J.~Jia, Hierarchical image saliency detection on extended cssd, IEEE transactions on pattern analysis and machine intelligence 38~(4) (2015) 717--729.

\bibitem{li2015visual}
G.~Li, Y.~Yu, Visual saliency based on multiscale deep features, in: Proceedings of the IEEE conference on computer vision and pattern recognition, 2015, pp. 5455--5463.

\bibitem{fan2020rethinking}
D.-P. Fan, Z.~Lin, Z.~Zhang, M.~Zhu, M.-M. Cheng, Rethinking rgb-d salient object detection: Models, data sets, and large-scale benchmarks, IEEE Transactions on neural networks and learning systems 32~(5) (2020) 2075--2089.

\bibitem{ji2022information}
Y.~Ji, P.~Jiang, J.~Shi, Y.~Guo, R.~Zhang, F.~Wang, Information-growth swin transformer network for image super-resolution, in: 2022 IEEE International Conference on Image Processing (ICIP), IEEE, 2022, pp. 3993--3997.

\bibitem{ji2023hyformer}
Y.~Ji, J.~Shi, Y.~Zhang, H.~Yang, Y.~Zong, L.~Xu, Hyformer: Hybrid grouping-aggregation transformer and wide-spanning cnn for hyperspectral image super-resolution, Remote Sensing 15~(17) (2023) 4131.

\bibitem{wang2024multiple}
T.~Wang, Z.~Zheng, Y.~Sun, C.~Yan, Y.~Yang, T.-S. Chua, Multiple-environment self-adaptive network for aerial-view geo-localization, Pattern Recognition 152 (2024) 110363.

\bibitem{wang2021pyramid}
W.~Wang, E.~Xie, X.~Li, D.-P. Fan, K.~Song, D.~Liang, T.~Lu, P.~Luo, L.~Shao, Pyramid vision transformer: A versatile backbone for dense prediction without convolutions, in: Proceedings of the IEEE/CVF international conference on computer vision, 2021, pp. 568--578.

\bibitem{yi2024gponet}
Y.~Yi, N.~Zhang, W.~Zhou, Y.~Shi, G.~Xie, J.~Wang, Gponet: A two-stream gated progressive optimization network for salient object detection, Pattern Recognition 150 (2024) 110330.

\bibitem{qin2019basnet}
X.~Qin, Z.~Zhang, C.~Huang, C.~Gao, M.~Dehghan, M.~Jagersand, Basnet: Boundary-aware salient object detection, in: Proceedings of the IEEE/CVF conference on computer vision and pattern recognition, 2019, pp. 7479--7489.

\bibitem{fan2017structure}
D.-P. Fan, M.-M. Cheng, Y.~Liu, T.~Li, A.~Borji, Structure-measure: A new way to evaluate foreground maps, in: Proceedings of the IEEE international conference on computer vision, 2017, pp. 4548--4557.

\bibitem{perazzi2012saliency}
F.~Perazzi, P.~Kr{\"a}henb{\"u}hl, Y.~Pritch, A.~Hornung, Saliency filters: Contrast based filtering for salient region detection, in: 2012 IEEE conference on computer vision and pattern recognition, IEEE, 2012, pp. 733--740.

\bibitem{fan2018enhanced}
D.-P. Fan, C.~Gong, Y.~Cao, B.~Ren, M.-M. Cheng, A.~Borji, Enhanced-alignment measure for binary foreground map evaluation, arXiv preprint arXiv:1805.10421 (2018).

\bibitem{shen2024minet}
K.~Shen, X.~Zhou, Z.~Liu, Minet: Multiscale interactive network for real-time salient object detection of strip steel surface defects, IEEE Transactions on Industrial Informatics (2024).

\bibitem{shen2022fully}
K.~Shen, X.~Zhou, B.~Wan, R.~Shi, J.~Zhang, Fully squeezed multiscale inference network for fast and accurate saliency detection in optical remote-sensing images, IEEE Geoscience and Remote Sensing Letters 19 (2022) 1--5.

\bibitem{zhou2024admnet}
X.~Zhou, K.~Shen, Z.~Liu, Admnet: Attention-guided densely multi-scale network for lightweight salient object detection, IEEE Transactions on Multimedia (2024).

\bibitem{lee2022tracer}
M.~S. Lee, W.~Shin, S.~W. Han, Tracer: Extreme attention guided salient object tracing network (student abstract), in: Proceedings of the AAAI conference on artificial intelligence, Vol.~36, 2022, pp. 12993--12994.

\bibitem{li2023lightweight}
G.~Li, Z.~Liu, X.~Zhang, W.~Lin, Lightweight salient object detection in optical remote-sensing images via semantic matching and edge alignment, IEEE Transactions on Geoscience and Remote Sensing 61 (2023) 1--11.

\bibitem{gongyangli2022lightweight}
Z.~GongyangLi, Z.~Bai, W.~Lin, H.~Ling, Lightweight salient object detection in optical remote sensing images via feature correlation, IEEE Trans. Geosci. Remote Sens 60 (2022) 5617712.

\bibitem{cui2023autocorrelation}
W.~Cui, K.~Song, H.~Feng, X.~Jia, S.~Liu, Y.~Yan, Autocorrelation-aware aggregation network for salient object detection of strip steel surface defects, IEEE transactions on instrumentation and measurement 72 (2023) 1--12.

\bibitem{lu2024low}
X.~Lu, Y.~Yuan, X.~Liu, L.~Wang, X.~Zhou, Y.~Yang, Low-light salient object detection by learning to highlight the foreground objects, IEEE Transactions on Circuits and Systems for Video Technology (2024).

\bibitem{zeng2023adaptive}
X.~Zeng, M.~Xu, Y.~Hu, H.~Tang, Y.~Hu, L.~Nie, Adaptive edge-aware semantic interaction network for salient object detection in optical remote sensing images, IEEE Transactions on Geoscience and Remote Sensing 61 (2023) 1--16.

\bibitem{zhuge2022salient}
M.~Zhuge, D.-P. Fan, N.~Liu, D.~Zhang, D.~Xu, L.~Shao, Salient object detection via integrity learning, IEEE Transactions on Pattern Analysis and Machine Intelligence 45~(3) (2022) 3738--3752.

\bibitem{luo2024spatial}
H.~Luo, J.~Wang, B.~Liang, Spatial attention feedback iteration for lightweight salient object detection in optical remote sensing images, IEEE Journal of Selected Topics in Applied Earth Observations and Remote Sensing (2024).

\bibitem{liu2024msrmnet}
X.~Liu, L.~Wang, Msrmnet: Multi-scale skip residual and multi-mixed features network for salient object detection, Neural Networks 173 (2024) 106144.

\bibitem{liu2023tcgnet}
Y.~Liu, L.~Zhou, G.~Wu, S.~Xu, J.~Han, Tcgnet: Type-correlation guidance for salient object detection, IEEE Transactions on Intelligent Transportation Systems 25~(7) (2023) 6633--6644.

\bibitem{li2023salient}
G.~Li, Z.~Bai, Z.~Liu, X.~Zhang, H.~Ling, Salient object detection in optical remote sensing images driven by transformer, IEEE Transactions on Image Processing 32 (2023) 5257--5269.

\end{thebibliography}

\end{document}